\documentclass{article} % For LaTeX2e
\usepackage{iclr2024_conference,times}

\usepackage{bm}
\usepackage{graphicx}
\newcommand*{\scalemath}[2]{\scalebox{#1}{\mbox{\ensuremath{\displaystyle #2}}}}
\usepackage{amsmath,amssymb,amsfonts}
\usepackage{color}
\usepackage{amsthm}
\usepackage{hyperref}
\usepackage{subfigure}
\usepackage{mathrsfs}
\usepackage{enumitem}
\usepackage{bbm}
\usepackage{wrapfig}
\usepackage{makecell}
\usepackage{multirow}
\usepackage[ruled,vlined]{algorithm2e}

\usepackage{booktabs}
\usepackage{pifont}
\usepackage{colortbl}
\usepackage[font=small]{caption}
\definecolor{mygraylite}{gray}{.94}
\definecolor{mygray}{gray}{.89}
\definecolor{darkergreen}{RGB}{21, 152, 56}
\definecolor{amber}{rgb}{1.0, 0.75, 0.0}
\definecolor{darkseagreen}{rgb}{0.56, 0.74, 0.56}
\definecolor{darkblue}{rgb}{0, 0, 54.5}
\definecolor{darkorange}{rgb}{220,88,42}

\newcommand*{\modelname}{\text{SNIP}\@\xspace}

\newlist{myitemize2}{itemize}{4}
\setlist[myitemize2,1]{label={},leftmargin=0em}
\newlist{myitemize}{itemize}{4}
\setlist[myitemize,1]{label=\textbullet,leftmargin=1em}

\usepackage{hyperref}
\usepackage{url}

\title{\modelname: 
Bridging Mathematical Symbolic and\\ Numeric Realms with Unified Pre-training
}

\author{Kazem Meidani\thanks{Equal Contribution. Contact email: mmeidani@andrew.cmu.edu} {}$^{ \ 1}$
,  Parshin Shojaee$^{* \ 2}$
\textbf{, Chandan K. Reddy}$^{ \ 2}$
\textbf{, Amir Barati Farimani}$^{\ 1,3}$\\
$^{1}$ Department of Mechanical Engineering, Carnegie Mellon University\\
$^{2}$  Department of Computer Science, Virginia Tech\\
$^{3}$  Machine Learning Department, Carnegie Mellon University\\
}

\iclrfinalcopy % Uncomment for camera-ready version, but NOT for submission.
\begin{document}

\maketitle
% \vspace{-0.5em}
\begin{abstract}
% \vspace{-0.5em}
% Symbolic mathematical expressions have historically been instrumental in representing complex patterns observed in natural systems. While deep learning has shown promise in understanding symbolic mathematics and extracting knowledge from numerical datasets, a gap persists in achieving mutual understanding between symbolic equations and their corresponding numerical values. This paper introduces the \textbf{S}ymbolic-\textbf{N}umeric \textbf{I}ntegrated \textbf{P}re-training (SNIP) model, a pioneering approach inspired by multi-modal pre-training models. SNIP is designed to bridge the divide between symbolic mathematics and numerical datasets, fostering a unified understanding 
% % in uni-modal settings 
% and enabling advanced multi-modal reasoning. Extensive experiments demonstrate SNIP's superior performance in various tasks, including understanding, property prediction, symbolic regression, and extrapolation. Our findings suggest that SNIP offers a groundbreaking approach to symbolic-numeric understanding and generation, setting a new benchmark for tasks involving multi-modal symbolic and numeric contexts.

%%%%%%%%%%%%%%%%%%%%%%
% VERSION 4 (Final Version)
In an era where symbolic mathematical equations are indispensable for modeling complex natural phenomena, scientific inquiry often involves collecting observations and translating them into mathematical expressions. Recently, deep learning has emerged as a powerful tool for extracting insights from data. However, existing models typically specialize in either numeric or symbolic domains, and are usually trained in a supervised manner tailored to specific tasks. This approach neglects the substantial benefits that could arise from a task-agnostic multi-modal
% unified
% and mutual 
understanding between symbolic equations and their numeric counterparts. To bridge the gap, we introduce \modelname, a Symbolic-Numeric Integrated Pre-training model, which employs contrastive learning between symbolic and numeric domains, enhancing their mutual similarities in the embeddings. By performing latent space analysis, we observe that \modelname provides cross-domain insights into the representations, revealing that symbolic supervision enhances the embeddings of numeric data and vice versa. We evaluate \modelname across diverse tasks, including symbolic-to-numeric mathematical property prediction and numeric-to-symbolic equation discovery, commonly known as symbolic regression. Results show that \modelname effectively transfers to various tasks, consistently outperforming fully supervised baselines and competing strongly with established task-specific methods, especially in the low data regime scenarios where available data is limited \footnote{Code and model are available at: \url{https://github.com/deep-symbolic-mathematics/Multimodal-Math-Pretraining}}. 
\end{abstract}

% \vspace{-1.2em}
\section{Introduction}
\label{sec:intro}
\vspace{-0.9em}
%%%%%%%%%%%%%%%%%%%%%%%%
%%%%%%%%%%%%%%%%%%%%%%%%
%First Paragraph
%%% Motivation 
% (Importance of Numeric Symbolic)
Throughout the history of science, symbolic mathematics has been unreasonably effective in representing natural phenomena \citep{math-language-1960}. Complex patterns of natural systems, represented as numeric data observations, can be elegantly abstracted using mathematical formulas. Mathematical symbolism has given us the language to describe, understand, and predict the natural world. The challenge of bridging the gap between the numeric observations and their mathematical symbolic representations has been a consistent focus in many scientific and engineering domains. Recognizing and exploring this connection is crucial, as it promises to drive advancements 
% and
% innovations 
in various fields.

\vspace{-0.3em}
%%%%%%%%%%%%%%%%%%%%%%%%
%%%%%%%%%%%%%%%%%%%%%%%%
%Second Paragraph
%%% Motivation 
% Previous Studies on Symbolic-Numeric and Lack of Unified Pre-training
In recent years, deep learning has demonstrated promising capabilities in learning from symbolic mathematics language as well as extracting insights from numeric data observations. Transformer models \citep{Attention-NeurIPS-2017}, in particular, have emerged as frontrunners in this field, effectively capturing patterns within mathematical expressions and solving complex tasks such as differential equations and function integration \citep{Lample-Deep-SR-ICLR-2020,NeuralIntegrator_AAAI2022}.
% , and mathematical theorem proving \citep{TheoremProving_NeurIPS2022}.
% Efforts have also been made to enhance the mathematical reasoning of language models, improving their performance in general math word problem solving \citep{math-reasoning-survey-2023, math-reasoning-2019}.
However, these models, while powerful, are not inherently designed to handle numeric data inputs. While some pre-trained symbolic regression models have been introduced to map numeric datasets to their governing mathematical expressions in a supervised manner \citep{Biggio-NeSymReS-ICML-2021, Kamienny-E2E-symbolic-NIPS-2022}, a gap still remains in developing a task-agnostic pre-training model capable of mutual understanding between the modalities of symbolic mathematical equations and their corresponding numeric data.

\vspace{-0.3em}
%%%%%%%%%%%%%%%%%%%%%%%%
%%%%%%%%%%%%%%%%%%%%%%%%
%Third Paragraph
% Previous Studies on Multi-modal Pre-training and its Potential for Our peoblem
Multi-modal pre-training models, exemplified by groundbreaking models like Contrastive Language-Image Pre-training (CLIP) \citep{CLIP-ICML2021}, have found a significant place in the deep learning landscape. CLIP has particularly set new standards in vision-language tasks, bridging the understanding between visual content and natural language descriptions. 
% \ps{This mutual comprehension across different data modalities has opened up opportunities for more intuitive and context-aware machine learning applications}. 
Expanding beyond traditional vision-language domains, recent studies have broadened multi-modal pre-training to include other modalities, such as audio and tabular data
% , leading to enhanced representations and performance 
\citep{opt-multimodal-audio-omniperception-2021, metatransformer-multimodal, multimodal-tabular-2023}. Additionally, previously untouched scientific domains, like molecular representation, are also benefiting from these advancements in multi-modal representations \citep{multimodal-molecule-momu-2022, MOFormer-multimodal-2023}. Nevertheless, the symbolic-numeric domain remains relatively unexplored.  Considering the foundational role of symbolic mathematics in science and the ubiquity of numeric data, an in-depth exploration of their mutual learning is not only timely but essential. 

\vspace{-0.3em}
%%%%%%%%%%%%%%%%%%%%%%%%
%%%%%%%%%%%%%%%%%%%%%%%%
%Third Paragraph
% What's our method?
% Motivated by these insights,
In this work, we present \textbf{S}ymbolic-\textbf{N}umeric \textbf{I}ntegrated \textbf{P}re-training (\textbf{\modelname}) to connect the two often distinct worlds of symbolic mathematical expressions and their corresponding numeric manifestations. The architecture of \modelname, depicted in Fig.~\ref{fig:snip}, incorporates dual Transformer encoders, with each encoder dedicated to learning the symbolic or numeric representations of mathematical functions. Subsequently, a task-agnostic contrastive objective is employed to enhance the similarity between (symbolic, numeric) pairs of data. The multi-modal pre-training of \modelname provides capabilities to understand and generate cross-modal content. Our experiments show that
\modelname achieves remarkable performance in cross-modal mathematical understanding and prediction tasks. Additionally, by combining \modelname with an equation generation decoder and exploiting its interpolatable latent space, we can effectively harness \modelname's mutual knowledge for the task of numeric-to-symbolic equation discovery (known as symbolic regression), achieving competitive results with state-of-the-art baselines. The major contributions of this work can be summarized as follows:

\vspace{-0.7em}
\begin{itemize}[leftmargin=*]
% \item We propose \modelname, a pioneering pre-training approach integrating symbolic and numeric mathematical domains through joint contrastive learning, yielding embeddings enriched by dual-domain insights.
% \item We propose \modelname, a pioneering pre-training method, bridging the mathematical symbolic and numeric domains, enabling enriched representations for both domains.
\item Proposing \modelname, a pioneering pre-training method that integrates mathematical symbolic and numeric domains through joint representation learning. This approach captures mutual relationships, delivering embeddings that are informed and enhanced by both domains.
\vspace{-0.2em}
\item Evaluating \modelname in cross-modal comprehension across different mathematical property prediction tasks. Our results indicate that \modelname
% , both in its original and fine-tuned variants, performs better than 
outperforms the fully supervised baselines, particularly in low data regime scenarios. 
 % with limited data
% By visualizing the latent embeddings, we also 
Visualizing the latent embeddings also confirms that \modelname's pre-trained representations reveal patterns linked to these cross-modal mathematical properties.
% \item We evaluate the cross-modal comprehension of \modelname representations through various mathematical property prediction tasks. Quantitative results show that \modelname, in both its original and fine-tuned variants, outperforms the task-specific fully supervised baseline, especially under few-shot scenarios where available data is limited. Qualitative findings on the visualization of embeddings also reveal that \modelname's pre-trained representations possess inherent patterns associated with cross-modal mathematical properties, resulting in the transferable superior performance for various cross-modal comprehension studies. 
% Our evaluations reveal that \modelname's pre-trained representations possess inherent patterns associated with cross-modal mathematical properties. FPr the task of predicting these cross-modal mathematical properties
% In both its original and fine-tuned variants, \modelname consistently outperforms the fully supervised baseline, especially under few-shot scenarios, demonstrating its ability to generalize when limited data is available.
\vspace{-0.2em}
\item Leveraging \modelname for numeric-to-symbolic equation generation task, commonly known as symbolic regression. In this task, after training an expression generation decoder on top of \modelname's numeric encoder, we exploit the high-quality semantic within \modelname's continuous and low-dimensional latent representations to perform latent space optimization with the objective of finding equations with balanced accuracy-complexity. Results show that \modelname achieves state-of-the-art performance on the well-known SRBench \citep{SRBench-Cava-NeurIPS-2021} benchmark.

% In this task, we utilize a novel two-step strategy that integrates \modelname with expression generation and latent space optimization within its interpolatable latent space. Results show that \modelname achieves state-of-the-art performance on the well-known SRBench benchmark.

% delivers performance on par with leading SR baselines on the well-known SRBench benchmark.
% achieves competitive performance with established SR baselines on the well-known SRBench benchmark.  
\vspace{-0.5em}
% \item We propose \modelname, a pioneering pre-training method that integrates mathematical symbolic and numeric domains through joint contrastive learning. This approach captures mutual relationships, delivering embeddings that are informed and enhanced by both domains.
% \vspace{-0.2em}
% \item Our study shows how \modelname's symbolic supervision elevates numeric embeddings, while its numeric supervision similarly strengthens symbolic representations.
% This interplay highlights the inherent bond between symbolic and numeric data, offering a powerful foundation for cross-domain mathematical studies.
% \vspace{-0.2em}
% \item Our comprehensive evaluations demonstrate \modelname's proficiency across diverse tasks such as mathematical property prediction and symbolic regression. \modelname not only outperforms supervised baselines but also competes strongly with task-specific SOTA methods, with notable efficacy in few-shot scenarios, underscoring its capability to generalize from limited data. 
% \vspace{-0.5em}
\end{itemize}
\vspace{-1.0em}
\section{Related Work}
\label{sec:related}
\vspace{-0.8em}
% \paragraph{Multimodal Pre-training Models.}
\noindent \textbf{Large-scale Pre-training.}
% A Foundation for Mutual Learning.
%% MMPTM: Literature, VLM, Contrastive Losses, Applications, importance, CLIP, Surveys, Other modalities
Our work is built upon an extensive body of research advocating the advantages of pre-training large models on large datasets \citep{survey-pretraining-foundation-2023,selfsupervised-multimodal-survey-2023}. 
% Early endeavors in pre-training predominantly focused on single modalities. Self-supervised learning (SSL) emerged as a cornerstone, where models were trained to predict parts of the data from other parts, effectively turning the data into its own supervision \citep{ssl-cookbook-2023}. This paradigm, though simple, proved transformative, especially in domains where labeled data was scarce.
Initially, pre-training was single-modal, with self-supervised learning (SSL) as a key paradigm that used data as its own supervision, especially useful where labeled data was limited \citep{ssl-cookbook-2023}. 
This paved the way for the emergence of multi-modal pre-training, where
% \textcolor{red}{As the research community delved deeper, models became stronger across different modalities.}
% , it became evident that real-world data often exists in multiple modalities.} 
% So, multi-modal pre-training
% Multi-modal pre-training later emerged, where
models are trained to understand relationships across different modalities \citep{multimodal-pretrain-survey-2023}. Vision and language have traditionally played the two main characters of pre-training models. For instance, 
% Contrastive Language-Image Pre-training (CLIP)
CLIP \citep{CLIP-ICML2021}, ALIGN \citep{ALIGN-ICML-2021}, and FLAVA \citep{flava-multimodal-2022} utilize image-caption pairs to construct jointly learned embedding spaces.
% by employing dual encoders, one for images and another for text, and 
These models are trained to align the embeddings of corresponding image-caption pairs while distancing unrelated pairs. 
The success of multi-modal pre-training in vision and language spurred its adoption in other domains. For example, recent works have extended this approach to videos, audio, and even tabular data \citep{opt-multimodal-audio-omniperception-2021,m5product-multimodal-other-2022,multimodal-tabular-2023}. 
% \citep{opt-multimodal-audio-omniperception-2021,m5product-multimodal-other-2022,multimodal-tabular-2023,clamp-multimodal-music-2023}
Specialized scientific domains have also embraced this paradigm. For instance, different models have emerged to learn joint representations of molecules \citep{multimodal-molecule-momu-2022, MOFormer-multimodal-2023}. 
% models that learn joint representations of molecular structures and their properties have emerged
Our work introduces a fresh perspective, intertwining symbolic mathematics with numeric observations. To this end, we use multi-modal pre-training's potential to deepen the symbolic-numeric mutual understanding.

\vspace{-0.3em}
% \paragraph{Symbolic Mathematics Models.}
\noindent \textbf{Deep Symbolic Mathematics.}
%%% deep learning for symbolic mathematics, linear algebra, arithmetic in transformers, numbers in language models, mathematical reasoning in language models
% The confluence of deep learning and mathematical reasoning has ushered in a new era of symbolic computation capabilities \citep{math-reasoning-2019, math-reasoning-survey-2023}.
%In recent years
Recently, deep learning models have made significant performance in the field of mathematical reasoning \citep{math-reasoning-2019, math-reasoning-survey-2023}. 
The Transformer models, originally designed for NLP tasks \citep{Attention-NeurIPS-2017}, have been repurposed with remarkable success in the domain of symbolic mathematics. It has powered models that can integrate functions \citep{Lample-Deep-SR-ICLR-2020, NeuralIntegrator_AAAI2022}, prove mathematical theorems \citep{TheoremProving_NeurIPS2022}, and perform numerical calculations, such as arithmetic operations \citep{charton2022linear,jelassi2023length}. 
% and even perform intricate numerical calculations, such as arithmetic operations \citep{charton2022linear,jelassi2023length}. 
These achievements underscore the flexibility and potential of deep learning models in abstract reasoning.
Beyond pure symbolic reasoning, there is also a growing interest in supplementing these models with numerical knowledge for improved mathematical understanding. For example, recent works have studied to enhance language models with numeric representations, aiming to improve their skills
% prowess
in mathematical word problem-solving \citep{mathbert-2021,MWP-BERT-numeric-LM-2022, number-NLP-ACL-2021, injecting-numeric-LM-2020}. Some recent studies have also explored different strategies for tokenizing and encoding numeric data, such as using multi-hot or continuous representation of numbers \citep{Biggio-NeSymReS-ICML-2021, pmlr-v202-becker23a, golkar2023xval}. Our work contributes a new angle to this growing field by integrating symbolic and numeric understanding in a unified multi-moal pre-training framework. By doing so, we not only capture the abstract representations of mathematical symbolic concepts but also their tangible numeric behaviors.
% , offering a holistic view of the symbolic-numeric spectrum.
% , offering a comprehensive view of the symbolic-numeric spectrum.
% In recent years, deep learning models have made significant strides in the field of mathematical reasoning \citep{math-reasoning-2019, math-reasoning-survey-2023}. Transformers \citep{Attention-NeurIPS-2017} have been explored for symbolic mathematics tasks such as function integration \citep{Lample-Deep-SR-ICLR-2020}, theorem proving \citep{TheoremProving_NeurIPS2022}, and performing numerical calculations like basic arithmetic operations \citep{charton2022linear}. Recently, incorporation of numerical knowledge and numeric representations in pre-training has also been investigated to enhance the performance of language models in mathematical word problem solving \citep{mathbert-2021,MWP-BERT-numeric-LM-2022, number-NLP-ACL-2021, injecting-numeric-LM-2020}. Our work builds upon the successful representations of symbolic mathematics in these works, and learns a multi-modal representation of these symbolic mathematical equations and their numerical behavior. 
% \vspace{-1.0em}

\vspace{-0.3em}
\noindent \textbf{Symbolic Regression.}
%%% Symbolic Regression / Scientific Discovery / importance / categories of methods / recent advances and end2end methods
Symbolic regression (SR) concentrates on discovering mathematical expressions for complex systems and representing data patterns in interpretable symbolic form. It has broad implications in both science and engineering, facilitating the modeling of diverse physical phenomena \citep{Cranmer-cosmology-NeurIPS-2020, Rudy-PDE-Science-2017,Meidani-IP2-2023}. 
% materials-sym-2019
% Recent advancements have explored traditional Genetic Programming (GP) algorithms \citep{Schmidt-Lipson-2009, pysr-2023}, as well as neural network-based approaches leveraging reinforcement learning (RL) \citep{DSR-Petersen-ICLR-2021} and combinations of these approaches \citep{AI-Feynman-Science-2020, mundhenk-seeding-GP-NeurIPS-2021}.  
Genetic Programming (GP) algorithms laid the foundation for SR, offering methods to search the vast space of mathematical expressions \citep{Schmidt-Lipson-2009, pysr-2023}. 
The ascent of deep learning subsequently gave rise to neural network-centric methods
% , often enriched by reinforcement learning, 
to reinforce SR's representational capabilities \citep{DSR-Petersen-ICLR-2021}. 
Some pioneering works also combined the evolutionary strengths of GP with the adaptability of neural networks, 
% aiming for a synergistic approach to SR
aiming for a better SR search \citep{AI-Feynman-Science-2020, mundhenk-seeding-GP-NeurIPS-2021}. 
However, these methods often struggle with challenges such as computational intensity, limited semantic depth, and the necessity to reinitiate search for different datasets. 
Inspired by the success of pre-trained Transformers in NLP, recent works introduced pre-trained models for SR \citep{Biggio-NeSymReS-ICML-2021, Kamienny-E2E-symbolic-NIPS-2022,TPSR-ArXiv2023}, using synthetic data and pre-trained priors for equation generation. 
% These models rely on synthetic data and the power of large-scale pretrained priors to generate equations in a single forward pass.
Our multi-modal pre-trained model, \modelname, advances this research towards a more insightful SR direction, leveraging rich encodings that harmoniously bridge symbolic equations with their numeric counterparts.

\vspace{-0.8em}
\section{Pre-training}
\label{sec:pretmethod}
\begin{figure}[t]
\centering
\includegraphics[width=0.9\linewidth]{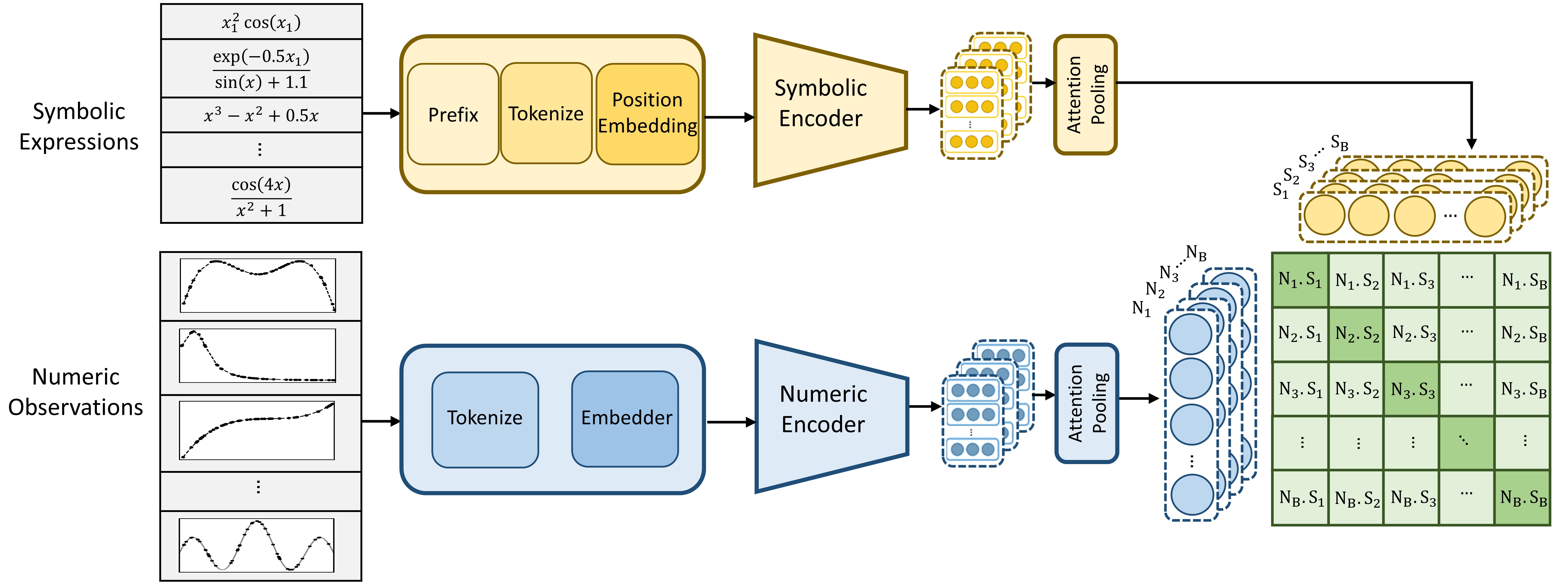}
% {Arxiv/fig_table_file/SNIP9.pdf}
% \captionsetup{font=footnotesize}
\caption{\small The \modelname Framework: A schematic representation of the dual-encoder pre-training scheme for mutual learning between symbolic equations and their numerical observations. Both symbolic and numeric encoders work in tandem, capturing the paired similarities and essence of their respective modalities. 
% Both encoders are trained using a contrastive objective, leveraging cosine similarity to ensure a coherent understanding between symbolic mathematical equations and their numerical counterparts. This design draws inspiration from the CLIP \citep{CLIP-ICML2021} model, adapting its principles for the unique challenges and potential of symbolic-numeric mutual learning.
% An overview of the pretraining scheme of \modelname with contrastive loss
\vspace{-0.5em}
}
\vspace{-0.5em}
\label{fig:snip}
\end{figure}

\vspace{-0.8em}
% In this section, we present the methodology behind 
%%%%%%%%%%
% \ps{repeated}
% \modelname introduces a unified pre-training framework tailored to enhance performance on symbolic-numeric understanding and generation tasks. 
%%%%%%%%%%
As depicted in Fig.~\ref{fig:snip}, the \modelname architecture comprises two Transformer encoders, each tailored for learning the symbolic or numeric representations of mathematical functions. These symbolic and numeric encoders are jointly trained with a task-agnostic contrastive learning objective to predict correct pairings within a batch of (symbolic, numeric) examples. During pre-training, \modelname receives synthetically created symbolic equations and their associated numeric data as inputs to the symbolic and numeric heads, respectively. In total, \modelname is pre-trained on approximately $60$ million synthetic paired examples.
% : one focuses on symbolic representations while the other targets numeric aspects of mathematical functions.
% incorporates dual transformer-based encoders, with each encoder dedicated to learning the symbolic or numeric representations of mathematical functions. 
% \modelname seamlessly combines a symbolic encoder with a numeric encoder. These encoders are jointly trained to predict correct pairings of a batch of (symbolic, numeric) training examples. 
% Throughout the pre-training phase, 
% the \modelname's unified pre-training framework is comprised of a symbolic encoder as well as a numeric encoder, which are jointly trained to predict correct pairings of a batch of (symbolic, numeric) training examples.
% The learned numeric/symbolic encoders are then utilized, based on the specific downstream task, to enhance performance over a diverse set of symbolic-numeric understanding and generation tasks. 
% unimodal and multi-modal tasks. 
% Depending on the downstream task, during inference, the learned numeric/symbolic encoders are used to facilitate performance on a wide set of unimodal or multi-modal tasks.
% \newpage
\vspace{-1.0em}
\subsection{Numeric Encoder}
\label{sec:numenc}
\vspace{-0.8em}
The numeric encoder's foundation is rooted in the recent advancements of Transformer models for encoding numeric observations into latent spaces \citep{Kamienny-E2E-symbolic-NIPS-2022,d-ascoli_icml22,Biggio-NeSymReS-ICML-2021}. In this framework, the numeric encoder—represented as $\mathcal{E}^{V}_{\theta}$—integrates an embedder, a multi-layer Transformer, and an attention pooling approach, to map numeric observations $(\bm{x},\bm{y})$ into a condensed latent vector $\bm{Z}_V$.
% In this framework, the numeric encoder—represented as $\mathcal{E}^{V}_{\theta}$—is a comprehensive module parameterized by $\theta$, encompassing an initial embedder, a multi-layer Transformer architecture, and an attention-based pooling stage. For a given numeric input observation, $(\bm{x},\bm{y})$,  this encoder yields a compact latent representation $\bm{Z}_V$.

\vspace{-0.3em}
\noindent \textbf{Tokenization.} Following \citep{charton2022linear,Kamienny-E2E-symbolic-NIPS-2022}, numeric inputs are tokenized using base-10 floating-point notation. They are rounded to four significant digits and subsequently represented as
% encoded into 
sequences of three tokens: sign, mantissa (0-9999 range), and exponent ($E$-$100$ to $E100$). For instance, the number $5.432$ is tokenized as $[+, 5432, E$-$3]$.

\vspace{-0.5em}
\noindent \textbf{Encoding.} Given a batch of $N$ numeric input points $(\bm{x},\bm{y})\in \mathbb{R}^{D+1}$, each is represented by $3(D+1)$ tokens. With increasing $D$ and $N$, the input sequence length grows, challenging the quadratic complexity of Transformers. To address this, we employ an embedder, as suggested by \citep{Kamienny-E2E-symbolic-NIPS-2022}, before the Transformer encoder. This embedder maps each input point to a unique embedding space. The resulting embeddings, with dimension $d_{\text{emb}}$, are then fed into the encoder.
% \noindent \textbf{Transformer Architecture:} 
For the numeric encoder, we utilize a multi-layer Transformer architecture \citep{Attention-NeurIPS-2017}.
% equipped with $16$ attention heads, $8$ layers, and an embedding dimension set to $512$.
Notably, due to the permutation invariance of the $N$ input points for each batch sample, we exclude positional embeddings, aligning with the approach in \citep{Biggio-NeSymReS-ICML-2021}. This encoder variant is denoted as $Enc^{V}$. The representation at its $l$-th layer is given by $\bm{V}_{l}=Enc^{V}_{l}(\bm{V}_{l-1})$, where $l$ ranges from 1 to $L_V$, and $L_V$ signifies number of layers within the numeric encoder.
% in the numeric encoder.

\vspace{-0.3em}
\noindent \textbf{Attention-based Distillation.} To distill the information from the Transformer's output into a compact representation for the whole sequence of observations, we employ an attention-based pooling mechanism, following \citep{attentivepooling1}. Let $\bm{\mathcal{A}}_{V}$ denote the attention weights, which are computed as: $\bm{\mathcal{A}}_{V} = \text{softmax}\left( \bm{W}_a \cdot \bm{V}^{T}_{L_V} \right)$, where $\bm{W}_a \in \mathbb{R}^{d_{\text{emb}}}$ is a learnable weight matrix, and we take the transpose of $\bm{V}_{L_V}\in \mathbb{R}^{N\times d_{\text{emb}}}$ to apply softmax along the sequence dimension $N$. The compact sequence-level representation, $\bm{Z}_{V}$, is then obtained by:
$\bm{Z}_{V} =  \bm{\mathcal{A}}_{V} \cdot \bm{V}_{L_V}$. This attention mechanism allows the model to focus on the most informative parts of the data points, effectively compressing the information into a fixed-size embedding.

\vspace{-1.0em}
\subsection{Symbolic Encoder}
\label{sec:symenc}
\vspace{-0.8em}
The symbolic encoder in our framework also draws inspiration from recent advancements in Transformer models for encoding symbolic mathematical functions, as demonstrated in works such as \citep{NeuralIntegrator_AAAI2022, Lample-Deep-SR-ICLR-2020}. Here, the symbolic encoder—denoted as $\mathcal{E}^{S}_{\psi}$—is a composite entity parameterized by $\psi$, encapsulating the embedder, a multi-layer Transformer, and attention-based pooling mechanisms. Given an input symbolic expression $f$, this encoder outputs a condensed representation $\bm{Z}_S$.
% Here, $\mathcal{E}^{S}_{\theta}$ refers to the whole symbolic head parameterized by $\theta$ which includes the embedder, transformer-based encoder, and the attention pooling weights.  

\vspace{-0.3em}
\noindent \textbf{Tokenization.} Mathematical expressions are tokenized by prefix order of their trees, following the principles outlined in \citep{Lample-Deep-SR-ICLR-2020}. This process employs self-contained tokens to represent operators, variables, and integers, while constants are encoded using the same methodology as discussed in Sec.~\ref{sec:numenc}, representing each with three tokens. In alignment with \citep{Lample-Deep-SR-ICLR-2020}, we use special tokens [$\langle\textit{BOS}\rangle$] and [$\langle\textit{EOS}\rangle$] to mark sequence start and end.
% we use special tokens [$\langle\textit{BOS}\rangle$] and [$\langle\textit{EOS}\rangle$] to signify the start and end of the sequence.

\vspace{-0.3em}
\noindent \textbf{Encoding.} Given a batch of symbolic expressions with $M$ tokens, each symbolic input is represented as $\bm{S}_0 = \left[ \bm{E}_{[\langle\textit{BOS}\rangle]}; \bm{E}_{t_1}; \ldots; \bm{E}_{t_M}; \bm{E}_{[\langle\textit{EOS}\rangle]} \right]+ \bm{S}^{pos}$, where $\bm{S}_0 \in \mathbb{R}^{(M+2)\times d_{\text{emb}}}$. Here, $\bm{E}$ refers to the embedding matrix, $t_i$ denotes the $i$-th token, $M$ signifies the number of tokens in the symbolic expression, $d_{\text{emb}}$ is the embedding dimension, and $\bm{S}^{pos}$ represents the positional embedding matrix. In the symbolic encoder, we use a Transformers model with the same architecture as in Sec.~\ref{sec:numenc}. This variant of the encoder, denoted as $Enc^{S}$, processes the symbolic inputs. The $l$-th layer representation is described as $\bm{S}_{l}=Enc^{S}_{l}(\bm{S}_{l-1})$, where $l$ varies from 1 to $L_S$, and $L_S$ indicates number of layers within the symbolic encoder.
% \noindent \textbf{Transformer Architecture.}

\vspace{-0.3em}
\noindent \textbf{Attention-based Distillation.} The symbolic encoder also employs attention-based pooling, as in Sec.~\ref{sec:numenc}. This mechanism computes weighted sums to distill information from the symbolic expression into a compact representation $\bm{Z}_{S} = \bm{\mathcal{A}}_{S} \cdot \bm{S}_{L_S}$, using attention weights $\bm{\mathcal{A}}_{S}$ through softmax along the symbolic sequence.

\vspace{-0.8em}
\subsection{Unified Pre-training Objective}
\label{sec:pretobj}
\vspace{-0.7em}
Our work introduces a multi-modal symbolic-numeric pre-training approach, \modelname, which 
% is inspired by the success of CLIP \citep{CLIP-ICML2021}. \modelname 
aims to facilitate a mutual understanding of both domains, enabling advanced cross-modal reasoning. 
%%%%%%%
% Unlike previous neuro-symbolic models, \modelname leverages both symbolic and numeric supervision to learn representations for each domain simultaneously. This unique approach opens up new possibilities in the realm of neuro-symbolic modeling, contributing to a wide range of tasks, including symbolic-numeric understanding and generation.

%%%%%%%%%%%%%
% USE THIS WRITING SOMEQHERE ELSE
% \noindent \textbf{Benefiting Diverse Tasks.} \modelname demonstrates its versatility by significantly improving performance on various tasks, such as property prediction, symbolic regression, and extrapolation. It achieves this by pre-training on multi-modal data and learning to correlate symbolic functions with numeric data. Remarkably, models using \modelname's pre-trained representations often outperform task-specific supervised models, highlighting the potential of multi-modal neuro-symbolic representations.

% \noindent \textbf{Scalability and Flexibility.} One of the key advantages of \modelname is its scalability. As a self-supervised model, it eliminates the need for labor-intensive crowd-sourced labeling commonly required for numeric data classification and prediction. Moreover, its ability to learn from multi-modal data fosters flexible and task-agnostic transfer learning, setting it apart from most uni-modal unsupervised or self-supervised learning approaches.
%%%%%%%%%%%%%

\vspace{-0.3em}
\noindent \textbf{Training Objective.} 
\modelname's pre-training objective is inspired by the joint training used in CLIP \citep{CLIP-ICML2021}. Incorporating both a numeric and symbolic encoder, the model optimizes a symmetric cross-entropy loss over similarity scores. It employs a contrastive loss (InfoNCE \citep{InfoNCE-2018} objective) to learn the correspondence between numeric and symbolic data pairs. Specifically, this approach learns to align embeddings of corresponding symbolic-numeric pairs while distancing unrelated pairs. The objective function can be defined as:
% It employs both a numeric encoder and a symbolic encoder to 
% \vspace{-0.2em}
\begin{equation}
\scalemath{0.92}{
\mathcal{L} = - \sum_{(v,s)\in B}{\big(\log{\text{NCE}(\bm{Z}_{S},\bm{Z}_{V})} + \log{\text{NCE}(\bm{Z}_{V},\bm{Z}_{S})} \big), }
}
\label{eq:loss}
\end{equation}

\vspace{-1.0em}
where $B$ represents the batch of (symbolic, numeric) data pairs, $\text{NCE}(\bm{Z}_{S},\bm{Z}_{V})$ and $\text{NCE}(\bm{Z}_{V},\bm{Z}_{S})$ denote the contrastive losses on symbolic-to-numeric and numeric-to-symbolic similarities, respectively.
% \vspace{-0.2em}
% \noindent \textbf{Contrastive Loss.} 
The symbolic-to-numeric contrastive loss, $\text{NCE}(\bm{Z}_{S},\bm{Z}_{V})$, is calculated as:
\vspace{-0.2em}
\begin{equation}
\scalemath{0.9}{
\text{NCE}(\bm{Z}_{S},\bm{Z}_{V}) = \frac{\exp{\big(\bm{Z}_{S} \cdot \bm{Z}_{V}^{+}\big)}}{ \sum_{{\bm{Z}\in\{\bm{Z}_{V}^{+},\bm{Z}_{V}^{-}\}}}{\exp{\left(\frac{\bm{Z}_{S} \cdot \bm{Z}}{\tau}\right)}}} 
}.
\label{eq:nceloss}
\end{equation}

\vspace{-1.0em}
where $\tau$ is temperature, $\bm{Z}_{V}^{+}$ represents positive \modelname numeric embeddings that overlap with \modelname symbolic embedding $\bm{Z}_{S}$, and $\bm{Z}_{V}^{-}$ are negative numeric embeddings implicitly formed by other numeric embeddings in the batch. A symmetric equivalent, $\text{NCE}(\bm{Z}_{V},\bm{Z}_{S})$, also defines the numeric-to-symbolic contrastive loss.
% \vspace{-0.1em}
More implementation details are provided in App.~\ref{sec:app-ptimp}.

\vspace{-0.9em}
\subsection{Pre-training Data}
\label{sec:pretrain_data}
\vspace{-0.8em}
In our \modelname approach, pre-training relies on a vast synthetic dataset comprising paired numeric and symbolic data. We follow the data generation mechanism in \citep{Kamienny-E2E-symbolic-NIPS-2022}, where each example consists of $N$ data points $(x,y)\in\mathbb{R}^{D+1}$ and a corresponding mathematical function $f$, where $y=f(x)$. Data generation proceeds in several steps, ensuring diverse and informative training examples. More details about each of the following steps are provided in App.~\ref{sec:app-pretraindata}.

\vspace{-0.3em}
\noindent \textbf{Sampling of functions.} We create random mathematical expressions using a process detailed in \citep{Kamienny-E2E-symbolic-NIPS-2022,Lample-Deep-SR-ICLR-2020}. This process involves selecting an input dimension $D$, determining the number of binary operators, constructing binary trees, assigning variables to leaf nodes, inserting unary operators, and applying random affine transformations. This method ensures a diverse set of functions for training. 
% For more details about each step, check App.~\ref{sec:app-pretraindata}.

\vspace{-0.3em}
\noindent \textbf{Sampling of datapoints.} After generating a function, we sample $N$ input points and find their corresponding target values. To maintain data quality, we follow guidelines from \citep{Kamienny-E2E-symbolic-NIPS-2022}, discarding samples with inputs outside the function's domain or exceptionally large output values. Our approach includes drawing inputs for each expression from various distributions, enhancing training diversity. The generation process of datapoints also involves selecting cluster weights and parameters, sampling input points for each cluster, and normalization along each dimension. To emphasize on the function's numeric behavior rather than the range of values, we also normalize the target values $\bm{y}$ between $(0,1)$. 

\vspace{-1.0em}
\section{Using \modelname for Cross-modal Property Prediction}
\label{sec:propmethod}
\vspace{-1.0em}
% \modelname-Guided Property Prediction
To evaluate \modelname's capability for cross-modal comprehension between symbolic and numeric domains, we conducted targeted experiments. These tests aimed to assess the model's aptitude for predicting specific numeric mathematical properties based on the symbolic inputs—a non-trivial task requiring mutual understanding of both domains.
% These tests aimed to assess the model's aptitude for predicting specific mathematical properties from one domain based on insights from the other—a non-trivial task requiring mutual understanding of both. 
For this purpose, we identified a set of mathematical properties; 
% encompassing both numeric and symbolic characteristics; 
details can be found in App.~\ref{sec:app-proppred}. In this section, we focus on two numeric properties for one-dimensional datasets: \textit{Non-Convexity Ratio (NCR)}, and Function \textit{Upwardness}. The \textit{NCR} approximates function convexity with values between \texttt{NCR=0} (fully convex) and \texttt{NCR=1} (fully concave). \textit{Upwardness} quantifies the function's directionality by assessing the segments where data increases within the training domain, 
% indicated by 
ranging from \texttt{UP=-1} for strictly decreasing functions to \texttt{UP=1} for increasing  ones. Due to space limitations, only results for \textit{NCR} and \textit{Upwardness} are discussed here. A complete list of properties with their detailed prediction results and corresponding chance levels, as well as their \modelname's pre-trained representations, are provided in App.~\ref{sec:app-proppred}.

\vspace{-0.8em}
\subsection{Models and Training}
\vspace{-0.7em}
% \vspace{-1.0em}
To assess property prediction on top of \modelname's embeddings, we employ a predictor head that passes these embeddings through a single-hidden-layer MLP to yield the predicted values. We adopt a Mean Squared Error (MSE) loss function for training on continuous properties. We consider three key training configurations to probe the efficacy of \modelname's learned representations:
\vspace{-0.8em}
\begin{itemize}[leftmargin=*]
\item \textbf{Supervised Model}: Utilizes the same encoder architecture as \modelname but initializes randomly.
% with random parameters.
\vspace{-0.4em}
\item \textbf{\modelname (frozen)}: Keeps the encoder weights fixed, training only the predictor head. 
\vspace{-0.4em}
\item \textbf{\modelname (finetuned)}: Initializes encoder from pretrained \modelname, allowing full updates during training.
\end{itemize}
\vspace{-0.8em}
% \vspace{-1.0em}
For a fair comparison, all model variants are trained on identical datasets comprising $10$K equations and subsequently tested on a distinct $1$K-equation evaluation dataset. These datasets are generated using the technique described in Sec.~\ref{sec:pretrain_data}.

\begin{figure}[t]
\vspace{2.0em}
% \RawFloats
\begin{minipage}{.63\textwidth}
\centering
\vspace{-3.5em}
\includegraphics[width=\linewidth]{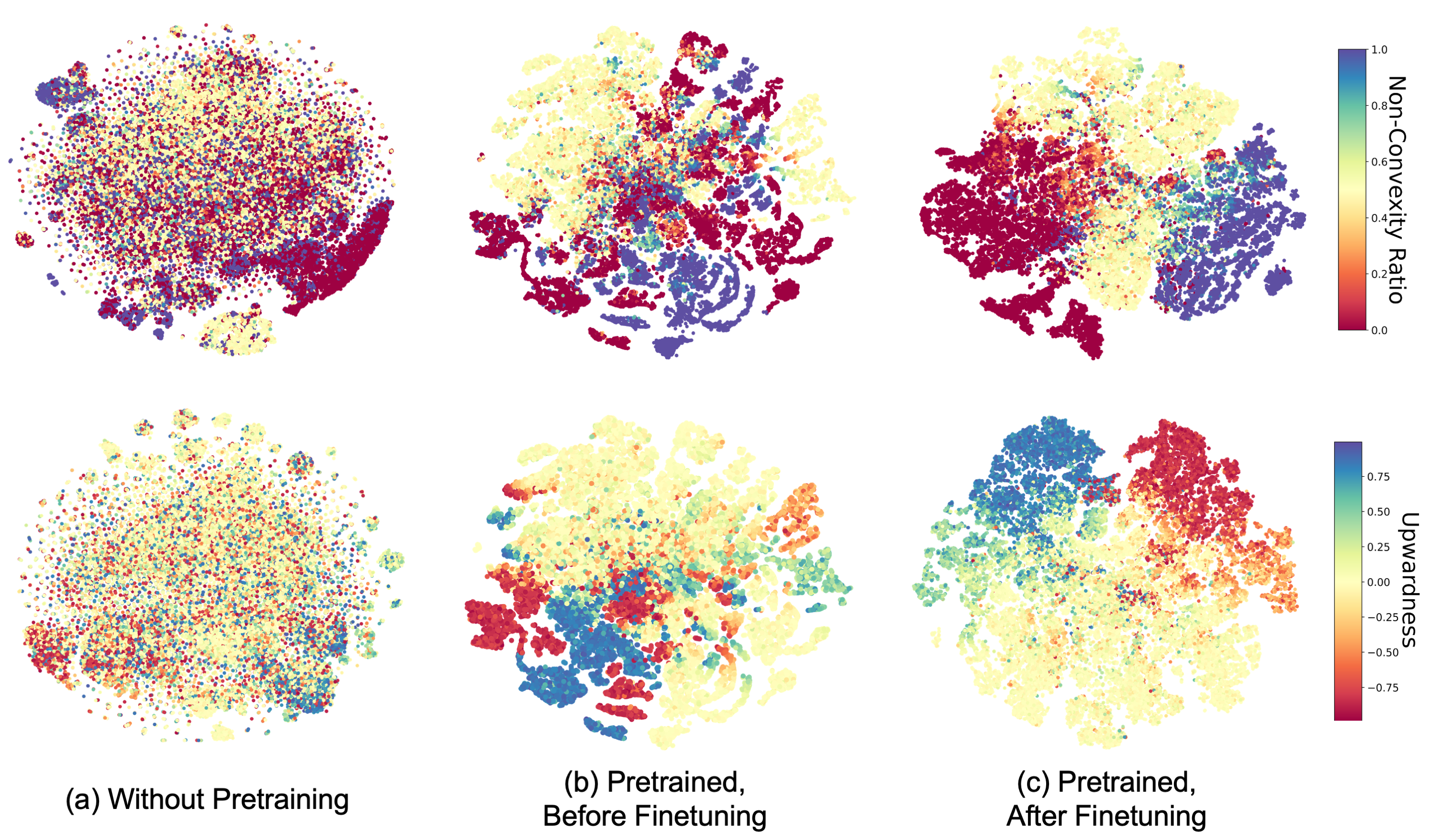}
\caption{\small 2D t-SNE representations of the encoded vectors across three model variants, colored for \textbf{(top)} Non-Convexity Ratio and \textbf{(bottom)} Function Upwardness prediction tasks.}
\label{fig:latentprop}
% \vspace{-0.5em}
\end{minipage}
\hfill
\begin{minipage}{.32\textwidth}
\centering
\vspace{-0.5em}
\includegraphics[width=\linewidth]{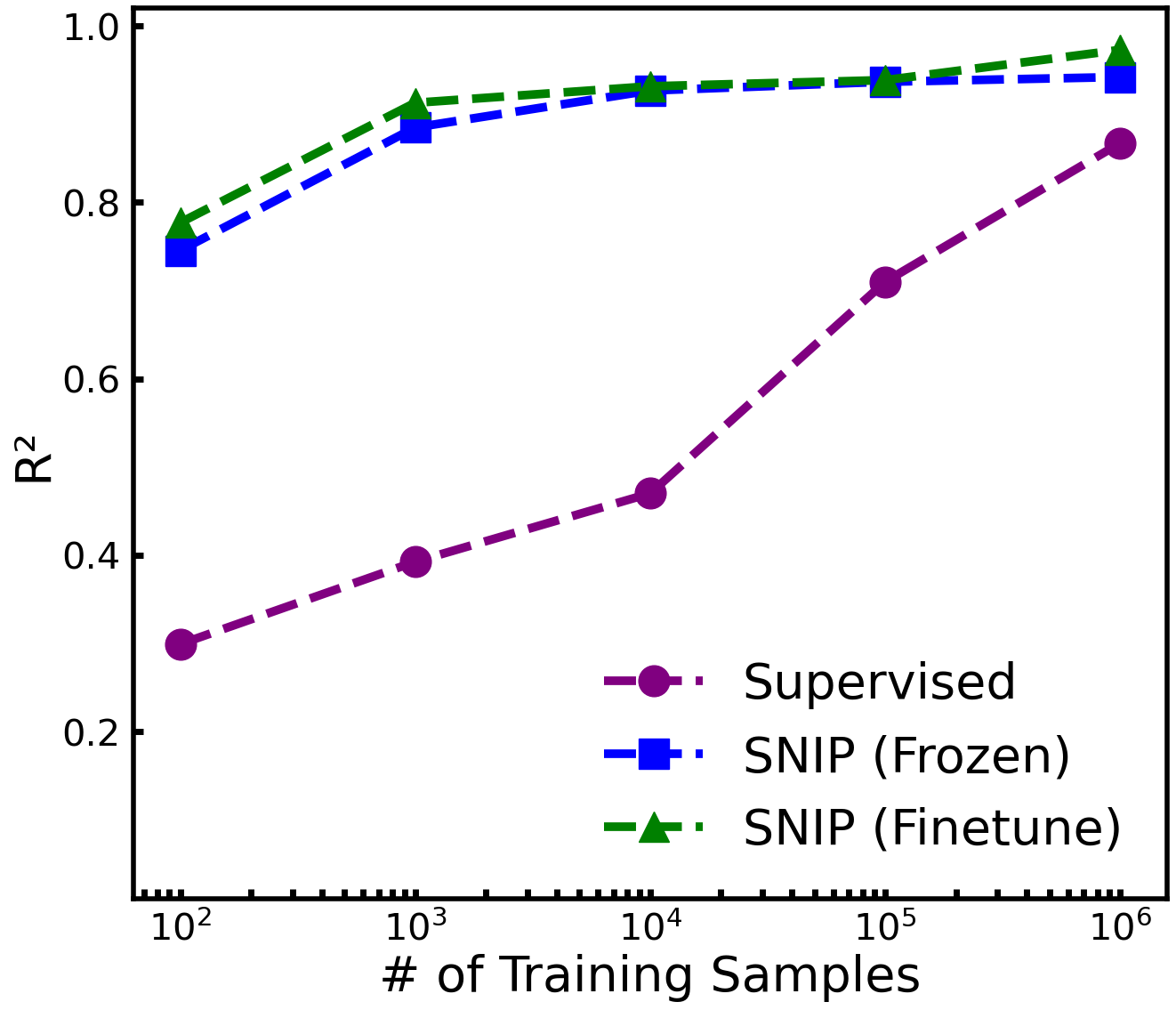}
\caption{\small $R^2$ scores for \textit{NCR} property prediction task vs. the number of training samples.}
\label{fig:fewshot}
% \vspace{-4.5em}
\vspace{-0.5em}
\end{minipage}
\vspace{-1.2em}
\end{figure}

\begin{wraptable}[7]{r}{0.55\linewidth}
% \vspace{-1.5em}
\centering
\vspace{-1.0em}
 \aboverulesep=0ex
 \belowrulesep=0ex
\renewcommand{\arraystretch}{1.0}
\caption{\small Results of using \modelname for property prediction.}
\label{tab:proppred}
\vspace{-0.5em}
\resizebox{\linewidth}{!}{
\begin{tabular}{lcccc}
\toprule
\multirow{2}{*}{Model}                                                 & \multicolumn{2}{c}{Non-Convexity Ratio} & \multicolumn{2}{c}{Upwardness} \\
\cmidrule(lr){2-3}
\cmidrule(lr){4-5}
   & $\downarrow$ NMSE  &   $\uparrow$ $Acc_{0.1}$  & $\downarrow$  NMSE & $\uparrow$ $Acc_{0.1}$ \\
   % &  &  \textcolor{blue}{$base=2.4\%$} & & \textcolor{blue}{$base=22.4\%$} \\
\midrule
Supervised & 0.5299 & 0.565 & 0.5356 & 0.563 \\

SNIP (frozen) & 0.0731  & 0.861 & 0.0540       & 0.847 \\
SNIP (finetuned) & \textbf{0.0683} & \textbf{0.921} & \textbf{0.0400} & \textbf{0.901} \\
\bottomrule
\end{tabular}
}
% \end{center}
% \end{table}
% \end{minipage}
\vspace{-0.7em}
\end{wraptable}

% \vspace{-0.5em}
\vspace{-0.7em}
\subsection{Results}
% \vspace{-0.5em}
% \vspace{-1.0em}
\vspace{-0.7em}
% \vspace{0.3em}
\mbox{}
\noindent \textbf{Quantitative Results.} Table \ref{tab:proppred} presents the Normalized Mean Squared Error (NMSE) and accuracy metric $Acc_{0.1}$ for all three models across the tasks of predicting \textit{NCR} and \textit{Upwardness}. Here, $Acc_{0.1}$ reflects the percentage of predictions within absolute tolerance $\tau$~=~0.1 of the true normalized values: $Acc_{\tau}=\frac{1}{N_{test}} \sum_{i}{\mathbbm{1}\left\{|\hat{p}_i - p_i| \leq \tau \right\}}$ where $p_i$ and $\hat{p}_i$ are the true and predicted property values for the $i$-th example.
% The NMSE chance level is $1$, while $Acc_{0.1}$ base (chance level) depends on the property's value distribution. 
Results reveal a significant gap in
performance between the purely supervised model and those benefiting from \modelname's prior knowledge.
This performance gap can be attributed to \modelname's pre-trained, semantically rich representations, enabling enhanced generalization to unseen functions. Additionally, fine-tuning the \modelname encoder results in marginal performance gains, indicating the model's capability to adapt to different downstream tasks.

% $R^2$ 
% and Normalized Mean Squared Error (NMSE) for  
% all three models across the tasks of predicting \textit{NCR} and \textit{Upwardness}. 
% 
% obtained by joint pretraining on a diverse range of samples,
% which empowers it to have superior generalization to data of unseen functions.
% extensive pretraining on a diverse range of samples, which empowers it to have superior generalization to unseen equations. 
% the table highlights a marginal enhancement in prediction performance when fine-tuning \modelname's encoder, indicating the adaptability of the encoded space to specific downstream tasks.

% Table \ref{tab:proppred} compares the testing

% $R^2$ and Normalized Mean Squared Error of the three models on both tasks of predicting NCR and Upwardness. The results indicate a significant gap between the supervised model's performance and the models that enjoy \modelname's prior knowledge. In fact, \modelname brings about much better generalization to unseen equations due to the large number of samples that it has been pretrained on. From the table, we also observe that finetuning the \modelname's encoder can also slightly improve the prediction performance by rearranging the encoded space based on the downstream task.  

\vspace{-0.3em}
\noindent \textbf{Qualitative Findings.} 
To delve deeper into the power of \modelname's representations, we compared its pre-finetuning and post-finetuning latent spaces against that of a supervised model lacking pretraining, using t-distributed Stochastic Neighbor Embedding (t-SNE) \citep{t-SNE-2008}.
% on a $50$K-sample dataset.
The visualizations are color-coded by the corresponding properties (Fig.~\ref{fig:latentprop}). Consistent with the quantitative outcomes, the supervised model's latent space, shown in Fig.~\ref{fig:latentprop}(a), exhibits limited structural coherence. In contrast, \modelname's latent space in Fig.~\ref{fig:latentprop}(b) shows pronounced clustering and distinct property trends. Notably, further fine-tuning of the encoder for these prediction tasks, depicted in Fig.~\ref{fig:latentprop}(c), results in a more structured latent space, marked by clearer linear trends in properties. This finding underscores \modelname's quantitative advantages and its flexibility in adapting to downstream tasks.
% with a more noticeable and linear property trend, reinforcing the observed quantitative advantages and the \modelname's easier adaptability to specific tasks.

\vspace{-0.3em}
\noindent \textbf{Low Data Regime Analysis.}
%%%%%%%%%%%%%%%%%%%%%%%%%%
We evaluated how training sample size influences the test \text{$R^2$ = $1-NMSE$} scores for predicting \textit{NCR}, assessing three model variants on a fixed $1$K-sample test set (Fig.~\ref{fig:fewshot}). 
In low data regime scenarios with as low as just $100$ training samples, the supervised model's score fell sharply to $0.292$, while both \modelname variants maintained scores above $0.745$.
% With $10$k samples, \modelname variants still outperform the supervised model with the roughly same outperformance gap.
Upon increasing the training sample size to $1$M, all models showed improvement; however, \modelname variants continued to lead. We observe that the supervised baseline model might approach \modelname's performance with more training data, which is reasonable, since this model is specialized only for the prediction of this property.
However, \modelname's value lies in its flexibility - the pre-trained representations can be efficiently adapted to new tasks.
% , unlike supervised baseline models trained from scratch for each task.
These results emphasize \modelname's superior generalization from limited data, underscoring the \modelname's rich semantic encodings.

\vspace{-1.0em}
\section{Using \modelname for Symbolic Regression}
\label{sec:srmethod}
\vspace{-1.0em}
% \modelname-Guided Symbolic Regression
% final_comment \km{Repeated} 
\modelname aims to synergize symbolic and numeric reasoning through mutual learning, offering enhanced capabilities for tasks that require both numeric-symbolic understanding and generation. A paramount task in this context is \textit{Symbolic Regression} (SR), which identifies interpretable symbolic equations to represent observed data. Essentially, SR transforms numeric observations into underlying mathematical expressions, thereby making it a numeric-to-symbolic generation task. The significance of SR extends to revealing functional relations between variables and offers an ideal benchmark for evaluating 
\modelname's pre-trained numeric representations. Recent advancements in SR leverage encoder-decoder Transformer frameworks \citep{Biggio-NeSymReS-ICML-2021, Kamienny-E2E-symbolic-NIPS-2022}. 
Therefore, to effectively undertake \modelname for SR, we perform the following two steps: First, training an expression generation decoder on top of the \modelname's numeric encoder for generating the symbolic functions. Second, conducting latent space optimization (LSO) within \modelname's interpolatable latent space, enriched by pre-training, to further enhance the equation generation.

\begin{figure}[t]
\centering
\includegraphics[width=0.7\linewidth]{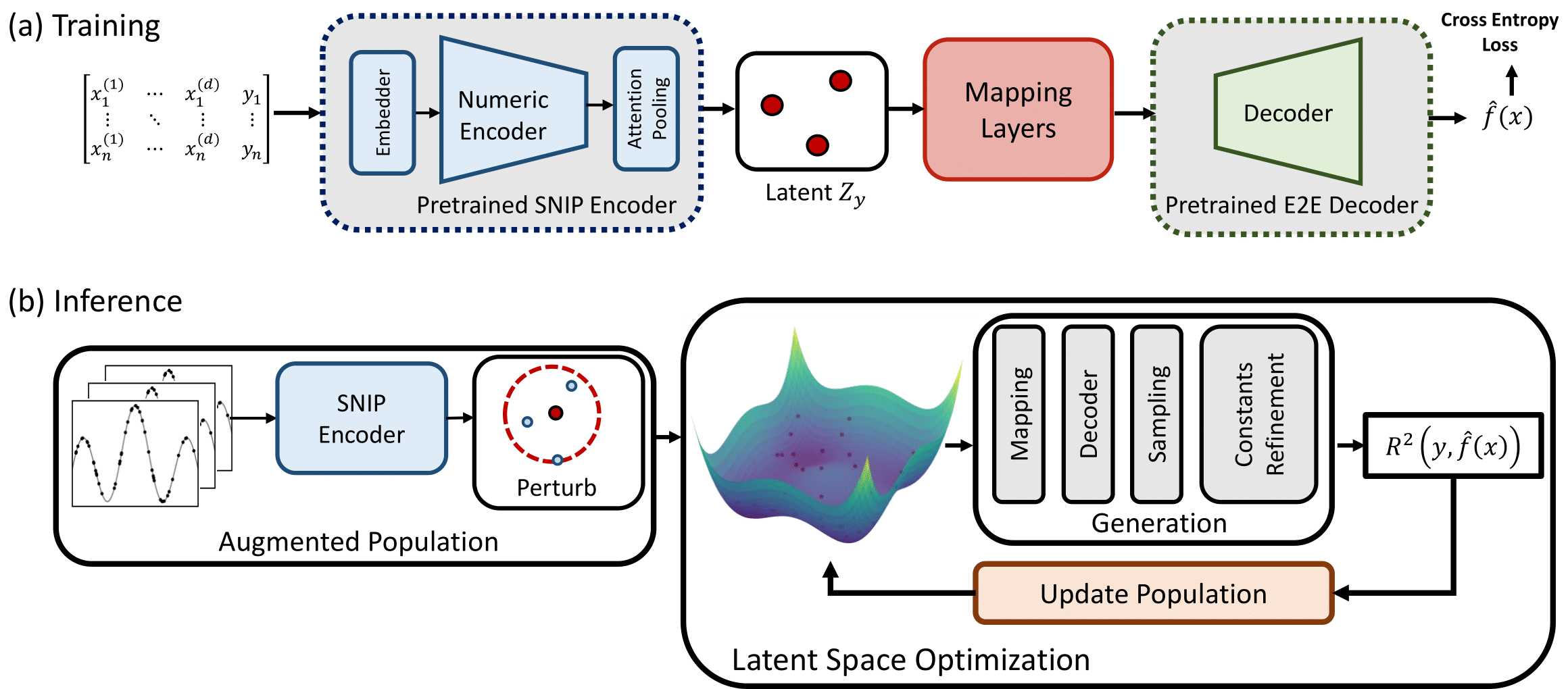}
% \captionsetup{font=footnotesize}
\caption{\small Using \modelname for Symbolic Regression: \textbf{(a) Training} includes adding an expression generation module atop \modelname's numeric encoder; \textbf{(b) Inference} aims to enhance expressions by optimizing within \modelname's interpolatable latent space.
% \textbf{(a) Training} an expression generation decoder on top of the \modelname's pre-trained numeric encoder; \textbf{(b) Inference} towards generating better expressions. During the inference phase, we leverage the interpolatability of \modelname's generative latent space and exploit latent space optimization for better generations. 
\vspace{-0.5em}
}
\vspace{-0.5em}
\label{fig:snipsrlso}
\end{figure}

\vspace{-0.8em}
\subsection{SR Model Architecture and Training}
\label{sec:srarch}
\vspace{-0.7em}
We build the SR model upon \modelname's numeric encoder $\mathcal{E}^{V}_{\theta}$ which transforms numeric data into semantically rich embeddings. On top of this encoder, we implement an expression generation module $\mathcal{G}_{\omega}$ that integrates an expression decoder $\mathcal{D}_{\phi}$ and a mapping network $g_{\gamma}$ to generate symbolic expressions: $\mathcal{G}_{\omega} = \mathcal{D}_{\phi} \circ g\left(\bm{Z}_V;\gamma\right)$. 
% \km{just checking this eq. Decoder (training) needs input seq. too. also why gamma is inside but phi and psi are subscript?} \ps{ there's no difference between inside and subscripts. both refer to the module parameters. This equation is equivalent to $\mathcal{D}_{\phi}(g_{\gamma}(\bm{Z}_V))$ mathematically}
% with the combined parameter set $\omega=\{\phi,\gamma\}$.

% \noindent \textbf{\modelname's Numeric Encoder.} 
% We utilize \modelname's numeric encoder, denoted as $\mathcal{E}^{V}_{\theta}$, to transform numerical observations of data into semantically-rich embeddings for subsequent modules in Symbolic Regression (SR).
%%%%%%%%%
% We utilize \modelname's numeric encoder $\mathcal{E}^{V}_{\theta}$ as the foundational layer for our Symbolic Regression (SR) framework. The encoder converts numerical data into semantically rich pre-trained embeddings, which serve as input for the subsequent modules. 
%%%%%%%%%
% This encoder has been pre-trained on large-scale paired (symbolic, numeric) samples and has demonstrated its capacity for capturing the essential features of symbolic-numeric data.

\vspace{-0.3em}
\noindent \textbf{Expression Decoder.}
% The expression ndecoder, denoted by $\mathcal{D}^{S}_{\phi}$, overlays after \modelname's numeric encoder. It employs a multi-layer Transformer architecture \citep{Biggio-NeSymReS-ICML-2021, Kamienny-E2E-symbolic-NIPS-2022} and is trained to convert numeric embeddings into symbolic expressions, aiming to minimize the divergence between the predicted $\hat{f}$ and the actual functions $f$.
%%%%%%%%%
To use \modelname for SR, we overlay an expression generation decoder $\mathcal{D}_{\phi}$, after \modelname's numeric encoder (shown in Fig.~\ref{fig:snipsrlso}(a)). This decoder, which utilizes a multi-layer Transformer \citep{Biggio-NeSymReS-ICML-2021, Kamienny-E2E-symbolic-NIPS-2022}, is trained to map numeric encodings into symbolic expressions, aiming to minimize the divergence between the predicted $\hat{f}$ and actual functions $f$.
%%%%%%%%%
% minimizing the discrepancy between generated symbolic expressions $\hat{f}$ and the ground-truth functions~$f$. 

% \mathcal{D}^{S}_{\phi} \circ g\left(\bm{Z}_V^i;\gamma\right)

% Our architecture consists of a numeric encoder $\mathcal{E}^{V}_{\theta}$, a mapping network  and an Expression Generation Decoder $\mathcal{D}^{S}_{\phi}$, jointly targeting an optimal symbolic function $f^*$ for approximating dataset $\mathcal{D} = \{\bm{x}_i, \bm{y}_i\}_{i=1}^N$. 

% To facilitate Symbolic Regression, we introduce an Expression Generation Decoder that operates on top of the pre-trained \modelname numeric encoder. This decoder is trained to generate symbolic mathematical expressions from the numeric representations encoded by \modelname. The decoder's training objective is to minimize the dissimilarity between the generated symbolic expressions and the true symbolic functions that correspond to the given numeric data. 

% In order to use \modelname for SR, we use the numeric encoder of \modelname. Then, we need to train an expression generation decoder on top of the \modelname's numeric encoder. This encoder-decoder framework helps us to map numeric observations into symbolic functions. Following existing works \citep{Biggio-NeSymReS-ICML-2021, Kamienny-E2E-symbolic-NIPS-2022}, we use transformer-based architecture for the decoder of this framework. 

\vspace{-0.3em}
\noindent \textbf{Mapping Network.}
Inspired by the ClipCap approach \citep{mokady2021clipcap} in the field of image captioning, which integrates CLIP's pre-trained image embeddings with GPT-2 pre-trained text generation model through a learnable mapping network, we adopt a similar strategy for SR. As shown in Fig.~\ref{fig:snipsrlso}(a), to facilitate integration with the E2E's \citep{Kamienny-E2E-symbolic-NIPS-2022} pre-trained SR decoder ($\mathcal{D}^{E2E}_{\phi}$), we introduce a learnable Mapping Network $g_{\gamma}$. This module translates \modelname's numeric embeddings $\bm{Z}_V$ into a compatible input for $\mathcal{D}^{E2E}_{\phi}$. 
% \km{maybe put D phi (without S) in other places and put E2E only here}. \ps{good point. modified.}
Specifically, $g:\mathbb{R}^{d_{emb}} \rightarrow \mathbb{R}^{M \times d_{emb}}$ reshapes \modelname embeddings into a sequence with maximum length $M$.
% , as the input prefix for $\mathcal{D}^{E2E}_{\phi}$. 
This approach lets us leverage the existing pre-trained SR decoder without the need for training from scratch.

\vspace{-0.3em}
\noindent \textbf{Training.}
The training objective is to minimize the token-matching cross-entropy loss $\mathcal{L}$ between the predicted $\hat{f}$ and ground-truth $f$ symbolic expressions:
% divergence between the generated symbolic expressions $\hat{f}$ and the true symbolic functions $f$. To achieve this, we employ a token-matching cross-entropy loss function with teacher forcing, defined as: 
$\mathcal{L}(\hat{f},f)=-\frac{1}{|f|}\sum_{j}{\log{P(\hat{t}_j|t_1,\ldots,t_{j-1}; \mathcal{G}_{\omega})}}$, where $P(\hat{t}_j|t_1,\ldots,t_{j-1}; \mathcal{G}_{\omega})$ is the conditional probability of the $j$-th token in $\hat{f}$, given the preceding true tokens. Here, the decoder is initialized from pre-trained weights \citep{Kamienny-E2E-symbolic-NIPS-2022} and trained jointly with the mapping network to learn numeric-to-symbolic expression generation. More details on the model designand training implementation can be found in App.~\ref{sec:app-sr}.

\vspace{-0.8em}
% \subsection{Latent Space Analysis}
% \subsection{Enhancing SR via \modelname's Semantic-rich Latents}
% Advancing SR with Semantic-rich Latent Insights of \modelname}
\subsection{Semantic Latent Insights for SR}
\vspace{-0.7em}
Traditional SR methods rely on searching within the vast equation landscape, dealing with the dual challenges of combinatorial complexity and limited prior knowledge \citep{operon-GP-2020,Schmidt-Lipson-2009}. 
Recent approaches incorporate deep learning to better navigate this space, integrating learned numeric-to-symbolic priors into the search process \citep{AI-Feynman-Science-2020,DSR-Petersen-ICLR-2021,Biggio-NeSymReS-ICML-2021,Kamienny-E2E-symbolic-NIPS-2022}. Yet, these are also often constrained by their reliance on the function search techniques at the decoding stage \citep{mundhenk-seeding-GP-NeurIPS-2021,DGSR-ICLR-2023,UDSR-NeurIPS-2022}, perpetuating the limitations. For example, in the Genetic Programming (GP) function search techniques, mutation and breeding steps across `winning' sub-expressions are prone to significant deviations in a function's numeric behavior. This emphasizes the necessity for a better search strategy attuned to the semantics of the function.
% volatile numerical behaviors with even minute changes. In other words, 
Recently, alternative strategies, like latent space learning of symbolic functions through Variational Autoencoders (VAEs) \citep{symbolic-VAE-2023,efficient-symbolic-VAE-2023}, trained exclusively for symbolic function reconstruction, do show promise but fall short by neglecting numeric behaviors essential for SR tasks.
% \km{essential for SR tasks?} \ps{yea synonyms}
% integral to SR tasks. 
% These latent spaces, trained exclusively for symbolic function reconstruction, lack nuanced numeric context essential for SR. 

\vspace{-0.2em}

\begin{wrapfigure}[15]{r}{0.32\linewidth}
% \centering
% \begin{figure}[!ht]
% \vspace{-0.5em}
\vspace{-1.5em}
\centering
\includegraphics[width=\linewidth]{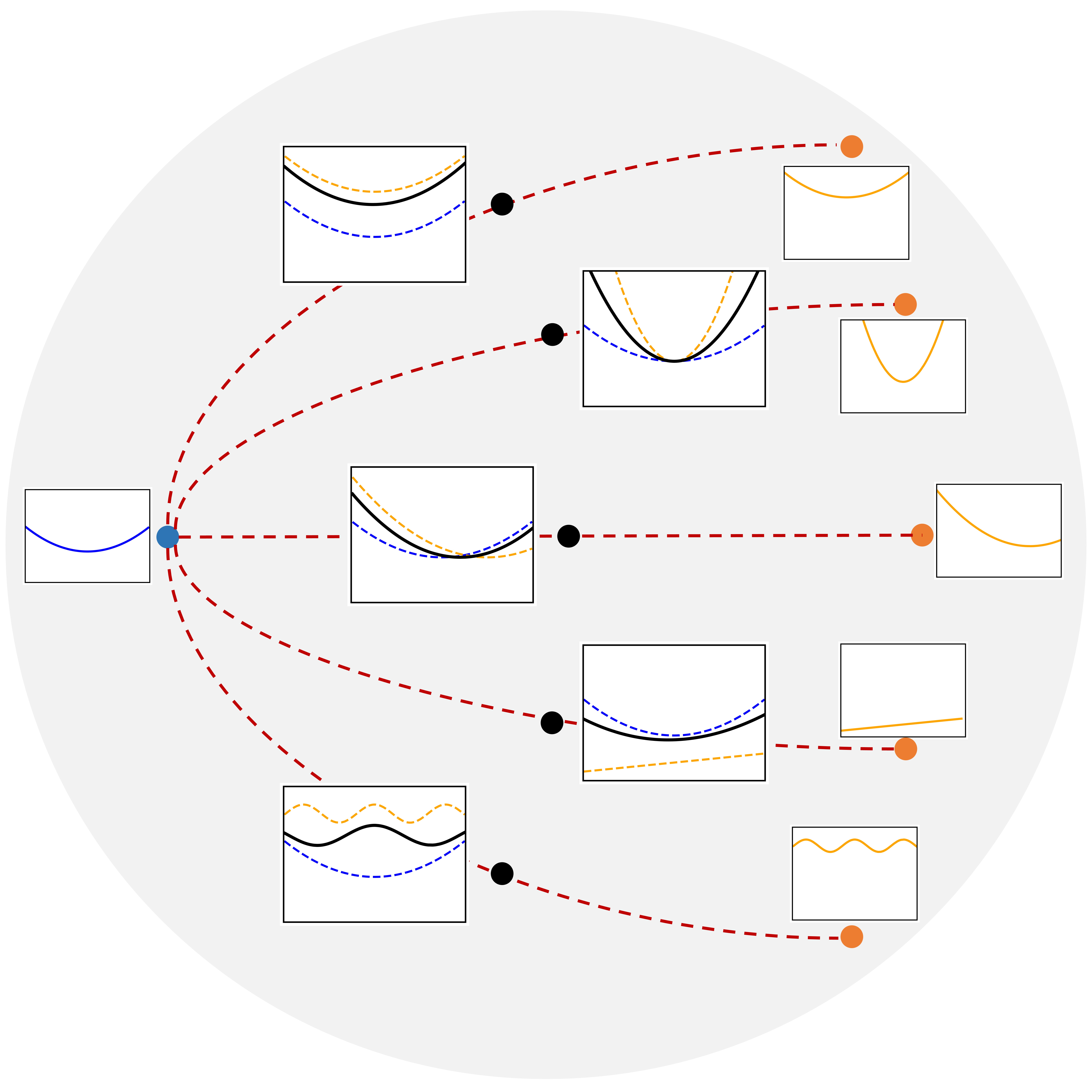}
% 
% \captionsetup{font=small}
% \vspace{-1.0em} 
\caption{\small Interpolatability of \modelname numeric latent space.
\vspace{-1.0em}
}
\label{fig:latentinterp}
\vspace{-0.7em}
% \end{figure}
\end{wrapfigure}
In contrast, \modelname offers a novel solution through a task-agnostic joint learning paradigm.
% final_comment \km{repeated} 
% to train on both symbolic functions and their corresponding numeric behaviors. 
This joint learning approach imprints the latent space with a wealth of integrated symbolic and numeric semantics that serve as a high-dimensional `semantic fingerprint' for various function behaviors and their inherent similarities. Therefore, unlike the latent space in \citep{symbolic-VAE-2023,efficient-symbolic-VAE-2023}, \modelname's task-agnostic latent space embodies a robust numeric-symbolic prior, providing an ideal landscape for SR search. By augmenting \modelname's numeric encoder with an expression generation decoder (as shown in Fig~\ref{fig:snipsrlso}), we can create a generative latent space—a crucial asset for the numeric-to-symbolic generation task of SR. 
% \vspace{-0.2em} 
Our empirical investigations on the generative latent space further enrich this narrative. 
% on top of the \modelname encoder 
% reveal an intriguing property—its intrinsic \textbf{\textit{interpolatability}}.
The innate \textbf{\textit{interpolatability} of SNIP's latent space}, as demonstrated in Fig.\ref{fig:latentinterp}, suggests a meaningful correlation between latent space representations and their corresponding numeric behaviors. 
In this figure, for a source function $\bm{Z}_V^s$ (blue curve) and a destination function $\bm{Z}_V^d$ (orange curves), we linearly interpolate within the numeric encoded vectors to obtain $\bm{Z}_V^{int}.$ This interpolated embedding is decoded into a symbolic function $\hat{f} = \mathcal{G}_{\omega} (\bm{Z}_V^{int.})$. Upon computing $\hat{f}$ over dataset $\bm{x}$, we find that \textit{the interpolated function exhibits a behavior that is semantically in between the source and destination functions.}
% The figure linearly interpolates between the numeric vectors of a source function $\bm{Z}_V^s$ (blue curve) and a destination function $\bm{Z}_V^d$ (orange curves) to produce $\bm{Z}V^{int.}$. Decoding this results in a symbolic function $\hat{f} = \mathcal{G}{\omega} (\bm{Z}_V^{int.})$. Plotting $\hat{f}$ on dataset $\bm{x}$, we find that interpolated functions bridge the behaviors of source and destination functions. 
% \km{In this figure, the numeric encoded vector of source function $\bm{Z}_V^s$ (blue curve) and the encoded vector of a destination function $\bm{Z}_V^d$ (each of the orange curves) are linearly interpolated to get $\bm{Z}_V^{int.}$. We then decode the interpolated embedding into symbolic function $\hat{f} = \mathcal{G}_{\omega} (\bm{Z}_V^{int.})$. Subsequently, we calculate the acquired function on the given dataset $\bm x$, and plot $\hat{f}$. We observe that for various shifts and changes in the function behavior, the interpolated functions exhibits a behavior that is semantically in between the source and destination functions.}
This is a significant advantage for nuanced search and explorations 
% and refinements 
during the symbolic discovery process. 
Moreover, the fixed dimension $d_{\text{emb}}$ of this space, which is substantially lower than the combinatorial optimization space of equations, 
% \km{I don't know if it's ok to compare dimensionality of a continuous space with combinatorial space of f...} 
streamlines the search process. Given these attributes, \modelname's generative latent space stands as a compelling candidate for a more effective approach to SR.

\vspace{-0.8em}
% \subsection{Latent Space Optimization}
% \subsection{Exploiting \modelname Latent Space for Optimized SR}
\subsection{\modelname Latent Space Optimization}
\vspace{-0.8em}
%% V1: Without mathematical notation 
As shown in Fig.~\ref{fig:latentinterp}, \modelname latent space interpolation shows a meaningful correlation with the functions' numeric pattern. This observation compels us to undertake a more comprehensive exploration of the latent space. Specifically, to fully harness the expressive capabilities of pre-trained \modelname embeddings in the context of SR, we employ Latent Space Optimization (LSO) as outlined in Fig.~\ref{fig:snipsrlso}(b). This optimization process involves a stochastic search over latent space $\bm{Z}_V$, with the objective of maximizing numerical fitness accuracy. To benefit from both prior knowledge of pre-trained model and capabilities of search method, we initialize the search population by augmenting the given dataset into a partitioned population $\mathcal{P} = \{\mathcal{P}_1, \mathcal{P}_2, \mathcal{P}_3\}$. Specifically, $\mathcal{P}_1$ contains encodings from random sub-samples, with size $n < N$ of the original data; $\mathcal{P}_2$ includes encodings from sampled inputs with their target values $\bm{y}$ perturbed by random Gaussian noise (perturb and then encode); and $\mathcal{P}_3$ includes perturbed encodings from a fixed sampled data (encode and then perturb). Each agent $p$ with representation $\bm{Z}_{V}^p$ is evaluated using a fitness function based on the $R^2$ fitting metric. Candidate symbolic functions are generated for each agent by feeding encodings to the expression generation module $\hat{f}_p = \mathcal{G}_{\omega}(\bm{Z}_V^p)$. The functions' constants are then refined using BFGS \citep{bfgs_flet87}, with a goal of optimizing the $R^2$ score against training data \citep{Kamienny-E2E-symbolic-NIPS-2022}.
% \vspace{-0.3em}
Then, updates to the latent population are carried out using a \textit{gradient-free optimizer}, which accommodates the non-differentiable characteristics of the function generation evaluation metrics.
% Then, latent population updates are conducted via a \textit{gradient-free optimizer} due to the non-differentiable nature of the evaluation scores for function generation. 
This latent optimization process \vspace{-0.2em} runs for $T$ iterations or until achieving a predefined $R^2_{\text{stop}}$ criterion. The optimal symbolic function $\hat{f}^*$ is then evaluated on a holdout test set. 
Overall, LSO leverages \modelname's rich latent space to efficiently transform symbolic regression's combinatorial search into continuous optimization of fitting performance. Details on the LSO algorithm and implementation are in App.~\ref{sec:app-sr}. An ablation study analyzing the impact of LSO and choice of optimization algorithm is also provided in App.~\ref{sec:app-srexps2}.

\vspace{-0.8em}
\subsection{Evaluation on SRBench}
\vspace{-0.8em}
\noindent \textbf{Datasets.}
% \vspace{-0.7em}
\modelname was assessed on PMLB datasets \citep{Olson2017PMLB} outlined in SRBench \citep{SRBench-Cava-NeurIPS-2021}, including: 119 \textit{Feynman} equations \citep{AI-Feynman-Science-2020}, 14 \textit{ODE-Strogatz} challenges \citep{LACAVA_strogatz2016}, and 57 \textit{Black-box} regression tasks without known underlying functions. For specifics on each dataset, refer to App.~\ref{sec:app-sr}. 
Leveraging the E2E's SR decoder \citep{Kamienny-E2E-symbolic-NIPS-2022} for our decoder initialization, which is trained for $D\leq10$, we similarly constrained \modelname's pre-training and evaluation to datasets with continuous features and dimensionality $D\leq10$. Also, since the range of target values $\bm{y}$ is important, especially for predicting the constants, we do not normalize $\bm{y}$ for this task. More details on the experiment settings are provided in App.~\ref{sec:app-sr}.

%% and implementation 
%%%%%%%%%
% We assess the prowess of \modelname on a collection of benchmark datasets sourced from the Penn Machine Learning Benchmark (PMLB) \cite{Olson2017PMLB}, as detailed in SRBench \cite{SRBench-Cava-NeurIPS-2021}. This assortment encompasses: ($i$) 119 equations from the Feynman Lectures on Physics database series \cite{AI-Feynman2_NeuRIPS2020}; ($ii$) 14 symbolic regression challenges from the ODE-Strogatz database \cite{LACAVA_strogatz2016}; and ($iii$) 57 Black-box regression problems devoid of known underlying functions. We limit the datasets to those with continuous features and input dimension $d\leq10$, as the \modelname is pre-trained with $d_{max}=10$ so that it can be aligned with the E2E's SR decoder \cite{Kamienny-E2E-symbolic-NIPS-2022} used as the initialization decoder for our SNIP-based SR model in Sec.~\ref{sec:srarch}. See App.~\ref{sec:app-srdatasets} for more detailed results.
%%%%%%%%%

\begin{figure}[t]
\centering
\includegraphics[width=\linewidth]{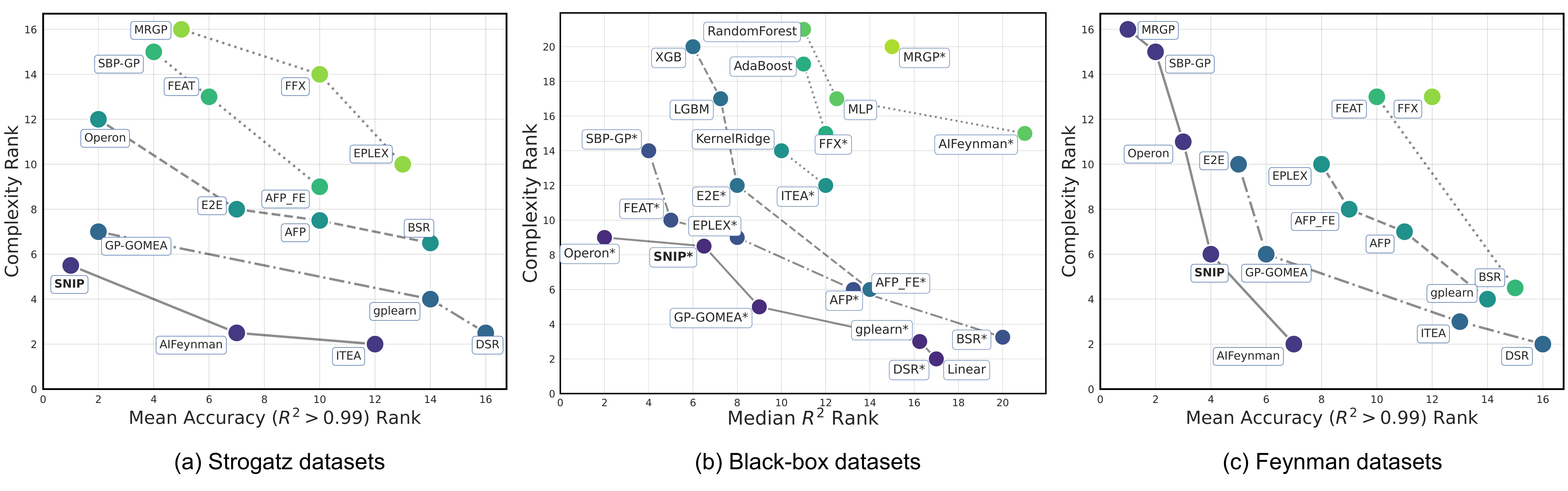}
% \captionsetup{font=footnotesize}
\caption{\small Pareto plots comparing $R^2$ and equation complexity of all methods across \textbf{SRBench datasets: (a)~\textit{Strogatz}}, \textbf{(b)~\textit{Black-box}}, and \textbf{(c)~\textit{Feynman}.} Using \modelname for SR yields strong fitting-complexity trade-off, evidenced by its first Pareto-front locating in all datasets. Here, each point depicts a method's median ranking within the data group, with lines/colors signifying Pareto dominance. The ``*'' marks SR methods in the \textit{Black-box} datasets. 
% in terms of the $R^2$ performance and identified equation complexity for \textbf{SRBench \textit{Black-box} dataset}, \textbf{(b) \textit{Feynman} dataset}, and \textbf{(c) \textit{Strogatz} dataset}. Our results by using pre-trained \modelname for the SR task show a great fitting-complexity trade-off, as it can be observed, \modelname is placed on the first Pareto-front on all three datasets. Connecting lines and colors denote Pareto-dominance rankings and ``*'' indicates SR methods (vs. ML methods) in \textit{Black-box} datasets. 
\vspace{-0.5em}
}
\vspace{-0.5em}
\label{fig:srbenchpareto}
\end{figure}

\vspace{-0.3em}
% \subsubsection{Results}
\noindent \textbf{Results.}
% \vspace{-0.7em}
Fig.~\ref{fig:srbenchpareto} illustrates \modelname's performance against the recent end-to-end (E2E) transformer SR model \citep{Kamienny-E2E-symbolic-NIPS-2022} and all the SRBench baselines. The Pareto plots exhibit rankings for \textit{Fitting Accuracy} against \textit{Model Complexity}. The model's accuracy is evaluated using $R^2$ and its complexity is evaluated as the number of nodes in the expression tree of the generated equation~\citep{SRBench-Cava-NeurIPS-2021}. Here, \modelname shows a strong accuracy-complexity balance, placing on the first Pareto-front across all datasets. 
On \textbf{\textit{Strogatz}} datasets, \modelname demonstrates top-tier accuracy of $0.928$, outperforming all the leading baselines. For \textbf{\textit{Black-box}} datasets, \modelname again shows competitive accuracy while achieving lower complexity ($47.52$) than the competitive Operon baseline ($64.95$). On \textbf{\textit{Feynman}} datasets, \modelname locates the Pareto frontier, offering better complexity than Operon ($31.63$ vs. $69.87$) and better accuracy than AIFeynman ($0.882$ vs. $0.798$) baselines. More detailed results on the SRBench datasets can be found in App.~\ref{sec:app-sr}.

% \newpage

% \section{Using \modelname for Extrapolation}
% \label{sec:extrapmethod}
% % \vsapce{-0.5em}

% \subsection{Model Architecture}

% \subsection{Training}

% \subsection{Results}

% \paragraph{Quantitative Results.}

% \paragraph{Qualitative Results.}
% \input{Arxiv/sections/discussion}
% \input{Arxiv/sections/experiments}

\vspace{-1.0em}
\section{Discussion and Conclusion}
\label{sec:conclusion}
\vspace{-0.8em}
%% Potentials
%% Limitations
%% future work
%%%%%%%%%%%%%%%%%%%%
% Results and Contributions
We introduced \modelname, a multi-modal symbolic-numeric pre-training model that learns how to associate the symbolic and numeric aspects of mathematical functions. We showed that \modelname exhibits remarkable capabilities in estimating cross-modal mathematical properties, particularly in low data regime scenarios, outperforming fully-supervised models. Also, by leveraging the latent space that \modelname constructs—capturing both functional behaviors and symbolic forms—the model demonstrates competitive performance in symbolic regression, even when compared to leading GP baselines.
% Also, by exploiting the \modelname's latent space learned by both functional behavior and symbolic forms, \modelname achieves competitive performance to the established baselines in symbolic regression. 
% In this work, we propose \modelname, a symbolic-numeric joint pre-training model that learns how to associate the symbolic and numeric aspects of mathematical functions. We showed that \modelname exhibits remarkable few-shot prediction capabilities in estimating mathematical properties, outperforming fully-supervised models. Also, by exploiting the \modelname's latent space 
% % optimizing in a latent space organized 
% mutually learned by both functional behavior and symbolic forms, \modelname achieves competitive performance to the established baselines in symbolic regression. 
%%%%%%%%%%%%%%%%%%%%%%%
%%%%%%%%%%%%%%%%%%%%%%%
%Limitation
%Parshinn Writing: 
While \modelname showcases robustness and versatility in integrating symbolic and numeric learning, it has notable limitations. It struggles with data patterns that cannot be clearly expressed as closed-form mathematical functions. Also, its performance is tied to the pre-defined data generation protocol, adopted from 
% prior works 
\citep{Lample-Deep-SR-ICLR-2020,Kamienny-E2E-symbolic-NIPS-2022}, which sets constraints on factors such as input dimensionality, and the vocabulary of mathematical operators. For example, the current protocol limits input dimensions to $D\leq10$ due to the exponential increase in expression complexity at higher dimensions. Exploring higher-dimensional settings is an interesting avenue for future research that would likely require significant updates to the data generation protocol.
Despite these limitations, \modelname has a wide range of capabilities, presenting a powerful tool in the intersection of symbolic and numeric mathematics.
% Looking forward, \modelname holds promise in diverse symbolic and numeric domains. 
%%%%%%%%
% Future work could explore its application in tasks ranging from symbolic function integration guided by numeric data, to numeric tasks like zero-shot extrapolation and super resolution enhanced by symbolic insights.
%%%%%%%%
Future research can focus on potential applications of \modelname, from using numeric guidance in symbolic-to-symbolic tasks such as function integration to using symbolic guidance for numeric-to-numeric tasks such as zero-shot extrapolation and super-resolution. Also, the \modelname's learned representations could serve as a foundation for innovative evaluation metrics of symbolic-numeric proximity, as well as efficient data and feature valuation.  

% offer a foundation 
%\textcolor{red}{(More potential applications are discussed in App. \ref{sec:app-discussion})}

\newpage

\bibliography{iclr2024_main.bbl}
\bibliographystyle{iclr2024_conference}

\newpage 

\appendix 
\section*{Appendix}

\vspace{-0.5em}
\section{Pre-training Data Details}
\label{sec:app-pretraindata}
\vspace{-0.5em}
We provide additional details regarding the pre-training data employed for pre-training \modelname. 
In our approach, \modelname is pre-trained on a large synthetic dataset of paired numeric and symbolic data, utilizing the data generation technique from \citep{Kamienny-E2E-symbolic-NIPS-2022}. Each example consists of a set of $N$ points $(\bm{x},y)\in\mathbb{R}^{D+1}$ and an associated mathematical function $f(\cdot)$, such that $y=f(\bm{x})$. These examples are generated by first sampling a function $f$, followed by sampling $N$ numeric input points ${\bm{x}_i;i=1,\ldots,N}\in\mathbb{R}^D$ from $f$, and then calculating the target value $y_i=f(\bm{x}_i)$.

\vspace{-0.5em}
\subsection{Sampling of functions}
\vspace{-0.5em}
% \noindent \textbf{Sampling of functions.}
To generate random functions $f$, we employ the strategy outlined in \citep{Kamienny-E2E-symbolic-NIPS-2022,Lample-Deep-SR-ICLR-2020}, building random trees with mathematical operators as nodes and variables/constants as leaves. This process includes: 

\noindent \textbf{Input Dimension Selection.} We begin by selecting the input dimension $D$ for the functions from a uniform distribution $\mathcal{U}(1, D_{max})$. This step ensures variability in the number of input variables.

\noindent \textbf{Binary Operator Quantity Selection.} Next, we determine the quantity of binary operators $b$ by sampling from $\mathcal{U}(D-1, D+b_{max})$ and selecting $b$ operators randomly from the set $\mathcal{U}({+,-,\times})$. This step introduces variability in the complexity of the generated functions.

\noindent \textbf{Tree Construction.} Using the chosen operators and input variables, we construct binary trees, simulating the mathematical function's structure. The construction process is performed following the method proposed in \citep{Kamienny-E2E-symbolic-NIPS-2022, Lample-Deep-SR-ICLR-2020}.

\noindent \textbf{Variable Assignment to Leaf Nodes.} Each leaf node in the binary tree corresponds to a variable, which is sampled from the set of available input variables ($x_d$ for $d=1,\ldots,D$). 

\noindent \textbf{Unary Operator Insertion.} Additionally, we introduce unary operators by selecting their quantity $u$ from $\mathcal{U}(0, u_{max})$ and randomly inserting them from a predefined set ($\mathcal{O}_u$) of unary operators where $\mathcal{O}_u=[ \rm{inv}, \rm{abs}, \rm{pow2}, \rm{pow3}, \rm{sqrt}, \sin, \cos, \tan, \arctan,\log, \exp]$.

\noindent \textbf{Affine Transformation.} To further diversify the functions, we apply random affine transformations to each variable ($x_d$) and unary operator ($u$). These transformations involve scaling ($a$) and shifting ($b$) by sampling values from $D_{\text{aff}}$. In other words, we replace $x_d$ with $ax_d+b$ and $u$ with $au+b$, where $(a,b)$ are samples from $D_{\text{aff}}$. This step enhances the variety of functions encountered during pre-training and ensures the model encounters a unique function each time, aiding in mitigating the risk of overfitting as well as memorization.

% $(i)$ selecting the desired function input dimension $D \sim \mathcal{U}(1,D_{max})$;
% $(ii)$ choosing the quantity of binary operators $b \sim \mathcal{U}(D-1,D+b_{max})$ and sampling $b$ operators from $\mathcal{U}(\{+,-,\times\})$; $(iii)$ constructing a binary tree with $b$ nodes using the sampling procedure from \cite{Kamienny-E2E-symbolic-NIPS-2022,Lample-Deep-SR-ICLR-2020}; $(iv)$ sampling one of the variables $x_d;d=1,\ldots,D$ for each tree leaf; $(v)$ selecting the number of unary operators $u \sim \mathcal{U}(0,u_{max})$, sampling $u$ operators from $O_u=[inv, abs, sqr, sqrt, sin, cos, tan, tan, atan,$ $log, exp]$, and randomly inserting them into the tree; $(vi)$ applying a random affine transformation to each variable $x_d$ and unary operator $u$, replacing $x_d$ with $ax_d+v$ and $u$ with $au+b$, where $(a,b)$ are samples from $D_{\text{aff}}$. As stated by \cite{Kamienny-E2E-symbolic-NIPS-2022}, this approach ensures the model encounters a unique function each time, aiding in pre-training and mitigating the risk of overfitting as well as memorization.

\vspace{-0.5em}
\subsection{Sampling of datapoints}
\vspace{-0.5em}
% \noindent \textbf{Sampling of datapoints.}
Once have generated a sample function $f$, we proceed to generate $N$ input points $x_i\in\mathbb{R}^D$ and calculate their corresponding target value $y_i=f(x_i)$. To maintain data quality and relevance, we follow the guidelines from \citep{Kamienny-E2E-symbolic-NIPS-2022}, which include:
\underline{\textit{Discarding and Restarting:}} If any input point $x_i$ falls outside the function's defined domain or if the target value $y_i$ exceeds $10^{100}$, we discard the sample function and restart the generation process. This ensures that the model learns meaningful and well-behaved functions.
\noindent \underline{\textit{Avoidance and Resampling:}} Avoidance and resampling of out-of-distribution $x_i$ values provide additional insights into $f$ as it allows the model to learn its domain. This practice aids the model in handling input variations.
\noindent \underline{\textit{Diverse Input Distributions:}} To expose the model to a broad spectrum of input data distributions, we draw input points from a mixture of distributions, such as uniform or Gaussian.  These distributions are centered around $k$ randomly chosen centroids, introducing diversity and challenging the model's adaptability.

The generation of input points involves the following steps:

\noindent \textbf{Cluster and Weight Selection.} We start by sampling the number of clusters $k$ from a uniform distribution $\mathcal{U}(1, k_{max})$. Additionally, we sample $k$ weights $\{w_j\sim\mathcal{U}(0, 1)\}_{j=1}^k$, which are normalized to $\sum_j{w_j}=1$.

\noindent \textbf{Cluster Parameters.} For each cluster, we sample a centroid $\mu_j\sim\mathcal{N}(0,1)^D$, a vector of variances $\sigma_j\sim\mathcal{U}(0,1)^D$, and a distribution shape $D_j$ from $\{\mathcal{N}, \mathcal{U}\}$ (Gaussian or uniform). These parameters define the characteristics of each cluster.

\noindent \textbf{Input Point Generation.} We sample $[w_jN]$ input points from the distribution $D_j(\mu_j, \sigma_j)$ for each cluster $j$. This sampling with different weights from different distributions ensures the sampling of a diverse set of input points with varying characteristics.

\noindent \textbf{Normalization.} Finally, all generated input points are concatenated and normalized by subtracting the mean and dividing by the standard deviation along each dimension.

% As recommended by \cite{Kamienny-E2E-symbolic-NIPS-2022}, if any $x_i$ falls outside the function's defined domain or if any $y_i$ exceeds $10^{100}$, we discard and restart the process. The avoidance and resampling of out-of-distribution $x_i$ values provide additional insights into $f$ as it allows the model to learn its domain. To enhance the variety of input distributions encountered during training, we draw our inputs from a mix of distributions (uniform or Gaussian), centered around $k$ randomly chosen centroids. The input points are generated through a series of steps: $(i)$~sample a number of clusters $k \sim \mathcal{U}(1,k_{max})$ and $k$ weights $\{w_j\sim\mathcal{U}(0,1)\}_{j=1}^k$, which are then normalized to $\sum_j{w_j}=1$; $(ii)$~sample a centroid $\mu_j\sim\mathcal{N}(0,1)^D$, a vector of variances $\sigma_j\sim\mathcal{U}(0,1)^D$ and a distribution shape $D_j\in\{\mathcal{N},\mathcal{U}\}$ for each cluster; $(iii)$~sample $[w_jN]$ input points from $D_j(\mu_j,\sigma_j)$ for each cluster; $(iv)$~concatenate all points obtained and normalize them by subtracting the mean and dividing by the standard deviation along each dimension.

% \vspace{-0.5em}
% \section{Pre-trained Model Details}
% \label{sec:app-pretrainmodel}
% \vspace{-0.5em}

% In this section, we provide further details on the implementation of \modelname and the training procedure. 

\vspace{-0.5em}
\section{Pre-training Implementation Details}
\label{sec:app-ptimp}
\vspace{-0.5em}
\subsection{Model Design Details}
\paragraph{Numeric Encoder.} 
The numeric encoding mechanism of our \modelname closely follows the design presented by \citep{Kamienny-E2E-symbolic-NIPS-2022}, as highlighted in Sec.~\ref{sec:pretmethod}. Firstly, for each instance in a given batch, the encoder receives $N=200$ numeric input points, $(\bm{x},\bm{y})$, from a space $\mathbb{R}^{D+1}$. Each of these points is tokenized into a sequence of length $3(D+1)$. An embedding module maps these tokens into a dense representation with an embedding size of $d_{\text{emb}}=512$.
% Notably, to ensure consistent input lengths across different samples, the \ps{input dimension is padded to a maximum of $D_{max}=10$}.
The sequences are then processed in the embedder module by a 2-layer feedforward neural network. This network projects input points to the desired dimension, $d_{\text{emb}}$. The output from the embedder is passed to a Transformer encoder, a multi-layer architecture inspired by \citep{Attention-NeurIPS-2017}. Our specific implementation has 8 layers, utilizes 16 attention heads, and retains an embedding dimension of $512$. A defining characteristic of our task is the permutation invariance across the $N$ input points. To accommodate this, we've adopted the technique from \citep{Kamienny-E2E-symbolic-NIPS-2022}, omitting positional embeddings within the numeric Transformer encoder. In our design, this specialized encoder variant is termed $Enc^{V}$. The representation generated at the $l$-th layer of the encoder is represented as $\bm{V}_{l}$. The process can be summarized as $\bm{V}_{l}=Enc^{V}_{l}(\bm{V}_{l-1})$. Here, the index $l$ spans from 1 to $L_V$, where $L_V=8$ denotes our encoder's total layers. Post encoding, for each instance in the batch, the numeric encoder's sequence outputs, $\bm{V}_{L_V}\in \mathbb{R}^{N\times d_{\text{emb}}}$, are compressed into a representation for the whole sequence, $\bm{Z}_{V}\in\mathbb{R}^{d_{\text{emb}}}$. This representation captures the essence of the entire numeric sequence and is achieved through an attention-pooling mechanism, detailed in Sec.~\ref{sec:numenc}.

\paragraph{Symbolic Encoder.} 
Our \modelname's symbolic encoding component draws inspiration from the model used in \citep{Lample-Deep-SR-ICLR-2020}, as highlighted in Sec.~\ref{sec:pretmethod}. This encoder is designed to process mathematical symbolic expressions with a maximum length of $200$. These expressions encapsulate the true functional relationships underlying the numeric data fed to the numeric encoder. The expressions are tokenized using a prefix order tree traversal. We employ the vocabulary defined by \citep{Kamienny-E2E-symbolic-NIPS-2022}, crafted to comprehensively represent mathematical equations. It includes symbolic entities like variables and operators, along with numeric constants. Constants are tokenized into three parts, consistent with the tokenization method outlined in Sec.~\ref{sec:numenc}. Sequence boundaries are indicated with special tokens [$\langle\textit{BOS}\rangle$] and [$\langle\textit{EOS}\rangle$]. Tokens are transformed into dense vectors of dimension $d_{\text{emb}}=512$ using an embedder module. This module essentially functions as an embedding matrix for the employed vocabulary. To maintain uniform input lengths, sequences are padded to a maximum length of $M=200$ and then projected to the desired embedding dimension. This dimensionality is aligned with the numeric encoder's. The embedded sequences are processed through a Transformer encoder, characterized by its multi-layer architecture as described by \citep{Attention-NeurIPS-2017}. Similarly, our specific configuration for this encoder consists of 8 layers, utilizes 16 attention heads, and retains an embedding dimension of $512$. Contrary to the numeric encoder, the sequence order in symbolic expressions holds significance. Consequently, we are including positional embeddings into this Transformer encoder variant. We denote this encoder as $Enc^{S}$, and its layer-wise representations are articulated as $\bm{S}_{l}=Enc^{S}_{l}(\bm{S}_{l-1})$, iterating from layer 1 to the maximum layer $L_S=8$. Similar to the numeric encoder's approach, the symbolic encoder condenses its Transformer outputs $\bm{S}_{L_S}\in \mathbb{R}^{M\times d_{\text{emb}}}$ for each expression into a compact representation, $\bm{Z}_{S}\in\mathbb{R}^{d_{\text{emb}}}$. This aggregation leverages the attention-pooling technique detailed in Sec.~\ref{sec:symenc}.

\vspace{-0.5em}
\subsection{Training Details}
% \paragraph{Training.} 
\vspace{-0.5em}
Following the extraction of coarse representations from both symbolic and numeric encoders, our focus shifts to harmonizing the embeddings from these encoders. The aim is to closely align embeddings representing corresponding symbolic-numeric pairs, while ensuring a discernible distance between unrelated pairs.  As discussed in Sec.~\ref{sec:pretobj}, this alignment process leverages a symmetric cross-entropy loss calculated over similarity scores, with the specific approach being informed by a contrastive loss mechanism. This ensures effective learning of the correspondence between numeric and symbolic data pairs. Our optimization process is facilitated by the \texttt{Adam} optimizer, operating on a batch size of $B=256$ (symbolic, numeric) data pairs. The learning rate initiation is set at a low $10^{-7}$, which is then gradually warmed up to $4\times10^{-5}$ over an initial span of $100\text{K}$ steps. Subsequently, in line with the recommendations of \citep{Attention-NeurIPS-2017}, we apply an inverse square root decay based on the step count to adjust the learning rate. Our model undergoes training for a total of $\approx 220$ epochs, with each epoch comprising $1,000$ steps.  This translates to the processing of $256\times1\text{K}=256\text{K}$ (symbolic, numeric) pair samples for each epoch. Given the on-the-fly data generation mechanism, as highlighted in Sec.~\ref{sec:app-pretraindata}, the cumulative volume of data encountered during pre-training approximates a substantial $60$M (symbolic, numeric) pair samples. For training, we utilize $4$ GPUs, each equipped with $48$GB of memory. Given this configuration, the processing time for a single epoch is approximately two hours.

\vspace{-0.5em}
\section{Details of Using \modelname for Cross-Modal Property Prediction}
\label{sec:app-proppred}
\vspace{-0.5em}

\subsection{Properties Definition}
\label{sec:app-proppred-define}
\begin{figure}[t]
% \begin{wrapfigure}[22]{r}{0.50\linewidth}
% \vspace{-0.5em}
\centering
\includegraphics[width=0.70\linewidth]{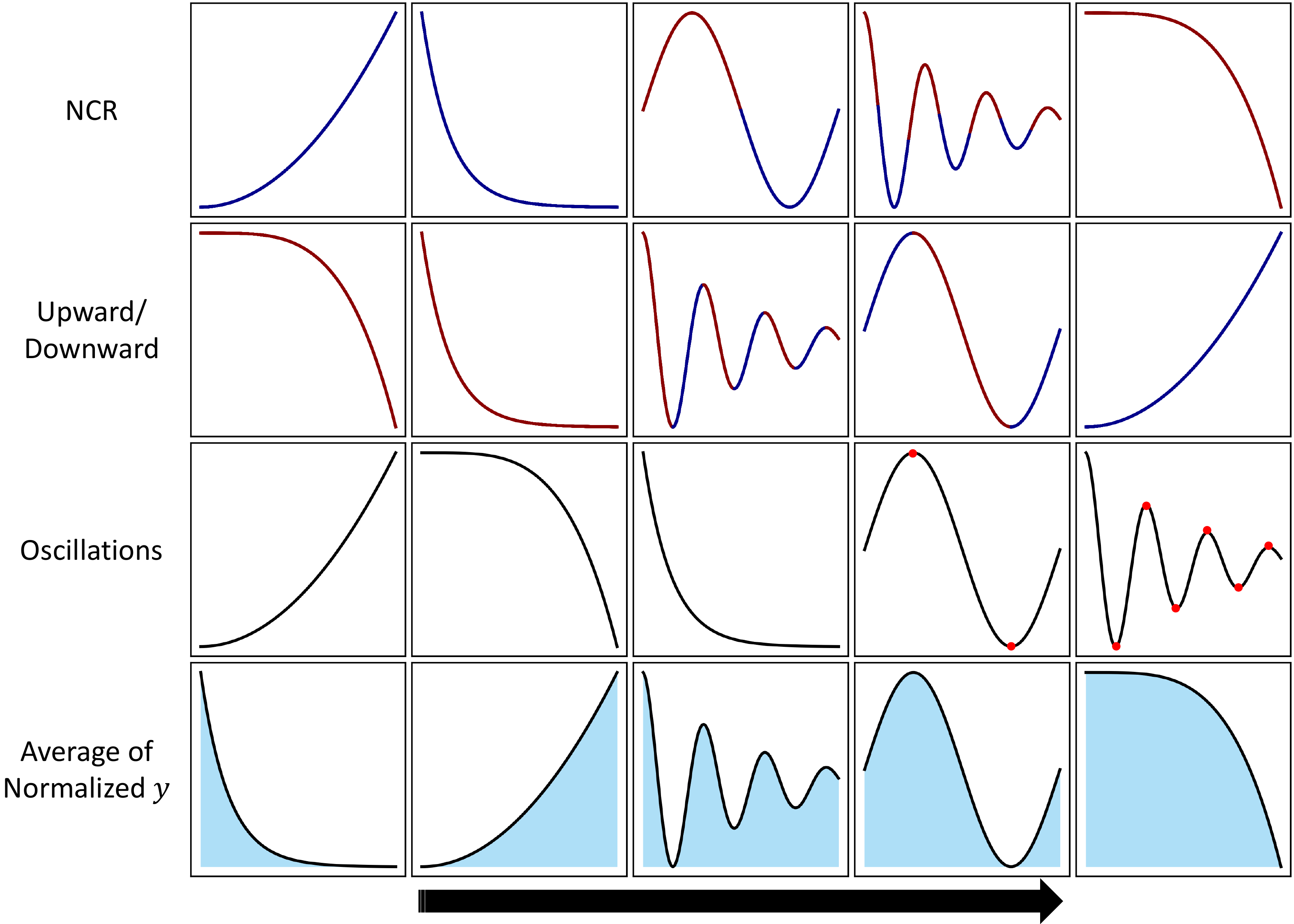}
\caption{\small Properties are qualitatively illustrated using five sample functions. Within each row, the plots are arranged according to their respective property values. Colors represent distinct function phases corresponding to the property (e.g., convexity vs. nonconvexity in the first row, upward vs. downward in the second row). Additionally, in the third row, red points highlight instances of change in the y-coordinate.
% Properties qualitatively depicted on five sample functions. In each row, the plots are sorted based on the corresponding values of the properties. The colors show the different phases of functions corresponding to the property (convexity vs. nonconvexity in the first row, upward vs. downward in second row) and the red points in the third row show the points of change in $y$. 
\vspace{-0.5em}
}
\label{fig:property_samples}
% \end{wrapfigure}
\end{figure}

In this section, we define the numeric mathematical properties that we use to evaluate the pre-trained \modelname model. The experiments include understanding and predicting numeric properties, i.e., properties that describe the behavior of numeric dataset, from symbolic forms of functions. The formal definitions of these properties are described in the following paragraphs and Fig.~\ref{fig:property_samples} qualitatively illustrates what each of the numeric properties represent. 

% , and 2) Understanding and predicting symbolic properties, i.e., properties that describe the symbolic functions, from numeric datasets.

\paragraph{Non-Convexity Ratio:}
% \vspace{-0.5em}
Non-Convexity Ratio (NCR) is defined to quantify the relative convexity (or non-convexity) of the functions as one of the properties depending on the numeric behavior of the functions. Hence, directly predicting this property from the symbolic form of the function is a complex task. To quantify the non-convexity ratio, we employ Jensen's inequality as a fundamental measure \citep{NCR-nonconvexity-2019}. In our approach, we focus on the one-dimensional equations with numeric dataset $ \{ \bm{x} ,\bm{y} \}$. Considering a function $f: \mathcal{D} \rightarrow \mathbb{R}$ where $\mathcal{D}$ is a convex subset of $\mathcal{R}$, $f$ is a convex function if $\forall x_1, x_2 \in \mathcal{D}$ and $\forall \lambda \in [0,1]$: 

\vspace{-0.5em}
$$ f(\lambda x_1 + (1-\lambda) x_2) \leq \lambda f(x_1) + (1-\lambda) f(x_2). $$

\vspace{-0.5em}
We rely on the training datasets with non-regularly sampled points to calculate the approximate NCR. To this end, we perform multiple trials to examine Jensen's inequality criterion. For each trial, we randomly select three data points $\{(x_i,f(x_i)), (x_j,f(x_j)), (x_k, f(x_k))\}$ which are sorted based on $x$ in ascending order. The convexity criterion holds on these points if

\vspace{-0.5em}
\begin{equation}
f(x_j) \leq \frac{{(x_k - x_j) \cdot f(x_i) + (x_j - x_i) \cdot f(x_k) }}{{x_k - x_i}} + \epsilon,
\label{eq:jensen}
\end{equation}

\vspace{-0.7em}
\noindent where $\epsilon$ is a very small number ($\epsilon=10^{-9}$) to avoid numerical precision errors. Therefore, for trial $t$, we define the success as 
\vspace{-0.5em}
$$ \xi_t = \begin{cases}
      1 & \text{if (\ref{eq:jensen}) holds,}\\
      0 & \text{otherwise.}\\
    \end{cases} $$

\vspace{-0.5em}
\noindent Finally, the non-convexity ratio (NCR) is computed over the total number of trials $T$ as

\vspace{-0.7em}
$$
NCR = 1 - \frac{1}{T} \sum_{t=1}^T {\xi_t}.
$$

\vspace{-0.5em}
Therefore, if a function is always convex over the range of training data points, \texttt{NCR=0}, and if it is always non-convex, it would have \texttt{NCR=1}. Functions that have both convex and non-convex sections in the range of $x$ will have \texttt{NCR} $\in (0,1)$.

\vspace{-0.5em}
\paragraph{Upwardness:} 
The `Upward/Downwardness' of a one-dimensional numeric dataset is defined to gauge the proportion of points within the training range where the function exhibits increasing or decreasing behavior. To compute this metric on the sorted dataset $\{\bm{x_{s}}, \bm{f(x_{s})}\}$, we examine every consecutive pair of points $\{x_i, x_{i+1}\}$ to determine if they demonstrate an upward or downward trend. We then define $u_{i}$ as follows:

% Upward/Downwardness of the numeric dataset in one-dimensional datasets is defined to measure the ratio of the points in the training range that the function has increasing or decreasing behavior. To quantify this metric on the sorted dataset $\{\bm{x_{s}} , \bm {f(x_{s})} \}$, we check for every consecutive points $\{ x_i, x_{i+1}\}$ if they have upward or downward behavior, and define $u_{i}$

\vspace{-0.7em}
$$ u_{i} = \begin{cases}
      1 & \text{if $f(x_{i+1}) > f(x_i) +\epsilon, $}\\
      -1 & \text{if $f(x_{i+1}) < f(x_i) -\epsilon, $}\\
      0 & \text{otherwise.}\\
    \end{cases} $$

\vspace{-0.7em}
Finally, the upwardness metric \texttt{UP} is computed as the average upwardness \texttt{UP} $ = \sum_{i=1}^{N-1} {u_i}$, where $N$ is the number of points in the dataset. Therefore, if a function is monotonically increasing the range of $x$ in training points, the upwardness measure is $1$, and if it is monotonically decreasing, the metric will be $-1$. Functions that have both sections in the range of $x$ will have \texttt{UP} $\in (-1,1)$.

\vspace{-0.5em}
\paragraph{Oscillation}
For this metric, we aim to quantify the degree of oscillatory behavior exhibited by the numeric data. This is approximated by counting the instances where the direction of $y$ changes. Determining the direction of data points follows a similar process to that of the upwardness metric for each consecutive pair. Thus, we tally the occurrences of direction changes while traversing the sorted dataset. Due to the potential variation in the number of changes, we opt for a logarithmic scale to color the plots.

% For this metric, we quantify the extent of oscillatory behavior that the numeric data shows. We approximate this metric by simply counting the number of times that the direction of $y$ are changing. The direction of the data points can be measured similar to the upwardness metric for every consecutive points. Therefore, we can count the number of times that the direction changes while going over the sorted dataset. Since the number of changes can vary a lot, we use logarithmic scale to coloring the plots.

%% TODO ? Need algorithm / equation to show the metric formally? 

\vspace{-0.5em}
\paragraph{Average of Normalized $y$}
The overall behavior of the numeric data points $\{ \bm{x}, \bm{y} \}$ are better represented when the values of $y$ are scaled to a fixed range (here $(0,1)$), giving $\{ \bm{x}, \bm{Y} \}$. The average of the normalized values, $\bar{Y}$ can be a measure to distinguish different numeric behaviors, and it can roughly approximate the numerical integral of the normalized function in the defined range of training $\bm{x}$.

% To delve deeper into the power of \modelname's representations, we compared its pre-finetuning and post-finetuning latent spaces against that of a supervised model lacking pretraining, using t-distributed Stochastic Neighbor Embedding (t-SNE) on a $50$k-sample dataset.  The visualizations are color-coded by the corresponding properties (Fig.~\ref{fig:latentprop}).

\subsection{Additional Quantitative Results of Cross-modal Property Prediction}
\label{sec:app-proppred-results}

\paragraph{Evaluation Metrics Overview.} We continue to use \textbf{NMSE} as our primary regression metric, providing a standard comparison across different model variants for each cross-modal property prediction task. We also report results for the \textbf{Accuracy within Tolerance ($Acc_{\tau}$)} evaluation metric, reflecting how closely the predicted values align with the true values, within a specified tolerance level. To this end, we first normalize the true and predicted values for each property in the range of $(0,1)$ based on the range of true values. Subsequently, we calculate the accuracy over $N_{test}=1000$ test examples as $Acc_{\tau}=\frac{1}{N_{test}} \sum^{N_{test}}_{i=1}{\mathbbm{1}\left\{|\hat{p}_i - p_i| \leq \tau \right\}}$, where $p_i$ and $\hat{p}_i$ are the normalized true and predicted values of the property for the $i$-th example, respectively. Here, we consider an absolute tolerance $\tau=0.1$. 

\paragraph{Chance Level Baselines.}
For each metric, we establish a baseline or chance level to set a comparative standard to better show task difficulty. For NMSE, the chance level is at $NMSE=1$, representing a prediction that averages the property values without considering the input. The chance level baseline for $Acc_{0.1}$ is calculated based on the assumption that all predictions are equal to the mean (average) property value, $\hat{p}_i = \bar{p}.$

\paragraph{Detailed Results.}
To offer a more detailed perspective on the performance of \modelname for cross-modal property prediction, we delve into its performance across four fundamental properties: Non-Convexity Ratio (NCR), Upwardness, Average of $y$, and Oscillations. Table \ref{tab:proppred_all} showcases a thorough comparison of the results from different model variants on these specified mathematical properties. A key aspect of these experiments is the exclusive use of symbolic equations as input for all models, aligning with the cross-modal essence of the tasks. The numeric properties are then predicted for these symbolic inputs, demonstrating the models' ability to bridge symbolic and numeric domains. For consistency in our evaluations, all models were trained on uniform datasets, each consisting of $10$K equations, and then assessed using a separate set of $1$K equations for evaluation. These datasets were constructed following the methodology outlined in Sec.~\ref{sec:app-pretraindata}. It's imperative to highlight that, in the context of cross-modal property prediction, \modelname operates with the same quantity of labeled examples as the supervised baselines. However, a critical distinction lies in \modelname's pre-training phase, where it was not exposed to any labeled data. Instead, it engaged in a multi-modal unsupervised learning process, focusing on capturing mutual symbolic-numeric similarities in representations. The results presented in Table \ref{tab:proppred_all} demonstrate that \modelname, both in its original 'frozen' state and when finetuned, consistently surpasses the performance of supervised models across all evaluated properties. This superiority is evident in both metrics – $NMSE$ and $Acc_{0.1}$. The variation in chance levels across different properties highlights the unique challenges inherent to each property. This variance underscores the adaptability and robustness of the \modelname model in navigating the diverse landscape of cross-modal property prediction tasks.

\begin{table}[t]
\centering
\renewcommand{\arraystretch}{1.1}
\caption{Full results of cross-modal property prediction on four properties showcase \modelname's superiority over the supervised baseline.}
\label{tab:proppred_all}
\vspace{-0.5em}
\resizebox{\linewidth}{!}{
\begin{tabular}{lcccccccc}
\toprule
\multirow{2}{*}{Model} & \multicolumn{2}{c}{Non-Convexity Ratio} & \multicolumn{2}{c}{Upwardness} & \multicolumn{2}{c}{Normalized Average $y$} & \multicolumn{2}{c}{Log Oscillations} \\ 
\cmidrule(lr){2-3}
\cmidrule(lr){4-5}
\cmidrule(lr){6-7}
\cmidrule(lr){8-9}
% \hline
& $\downarrow$ NMSE & $\uparrow$ $Acc_{0.1}$ & $\downarrow$ NMSE & $\uparrow$ $Acc_{0.1}$ & $\downarrow$ NMSE & $\uparrow$ $Acc_{0.1}$ & $\downarrow$ NMSE & $\uparrow$ $Acc_{0.1}$ \\ 
% &  & $base = 2.4\%$ &  & $base = 22.4\%$ &  & $base = 52.4\%$ & & $base = 23.1\%$ \\ 
\midrule
Base  & 1.0000 & 2.4\% & 1.0000 & 22.4\% & 1.0000 & 52.4\% & 1.0000 & 23.1\% \\ 
% \hline
Supervised & 0.5299 & 56.5\% & 0.5356 & 56.3\% & 1.0406 & 49.3\% & 0.3079 & 75.2\% \\ 
% \hline
SNIP (frozen) & 0.0731 & 86.1\% & 0.0540 & 84.7\% & 0.4532 & 64.5\% & 0.0683 & 92.6\% \\ 
% \hline
SNIP (finetuned) & \textbf{0.0683} & \textbf{92.1\%} & \textbf{0.0400} & \textbf{90.1\%} & \textbf{0.4074} & \textbf{67.7\%} & \textbf{0.0581} & \textbf{92.6\%} \\ 
\bottomrule
\end{tabular}
}
\end{table}

% To provide a more comprehensive view of SNIP's capabilities for cross-modal property prediction, we include full results on four key properties: Non-Convexity Ratio (NCR), Upwardness, Average of $y$, and Oscillations. The Normalized Average reflects the mean of normalized target values $y$ and indicates the overall trend of the function. Log Oscillation Count approximates oscillatory behavior by tallying direction changes in $y$. Formal definitions are provided in Appendix C.

\subsection{Additional Qualitative Findings of Cross-modal Property Prediction}
\label{sec:app-proppred-latent}

\begin{figure}[t]
\centering
% \vspace{-1.5em}
\includegraphics[width=0.95\linewidth]{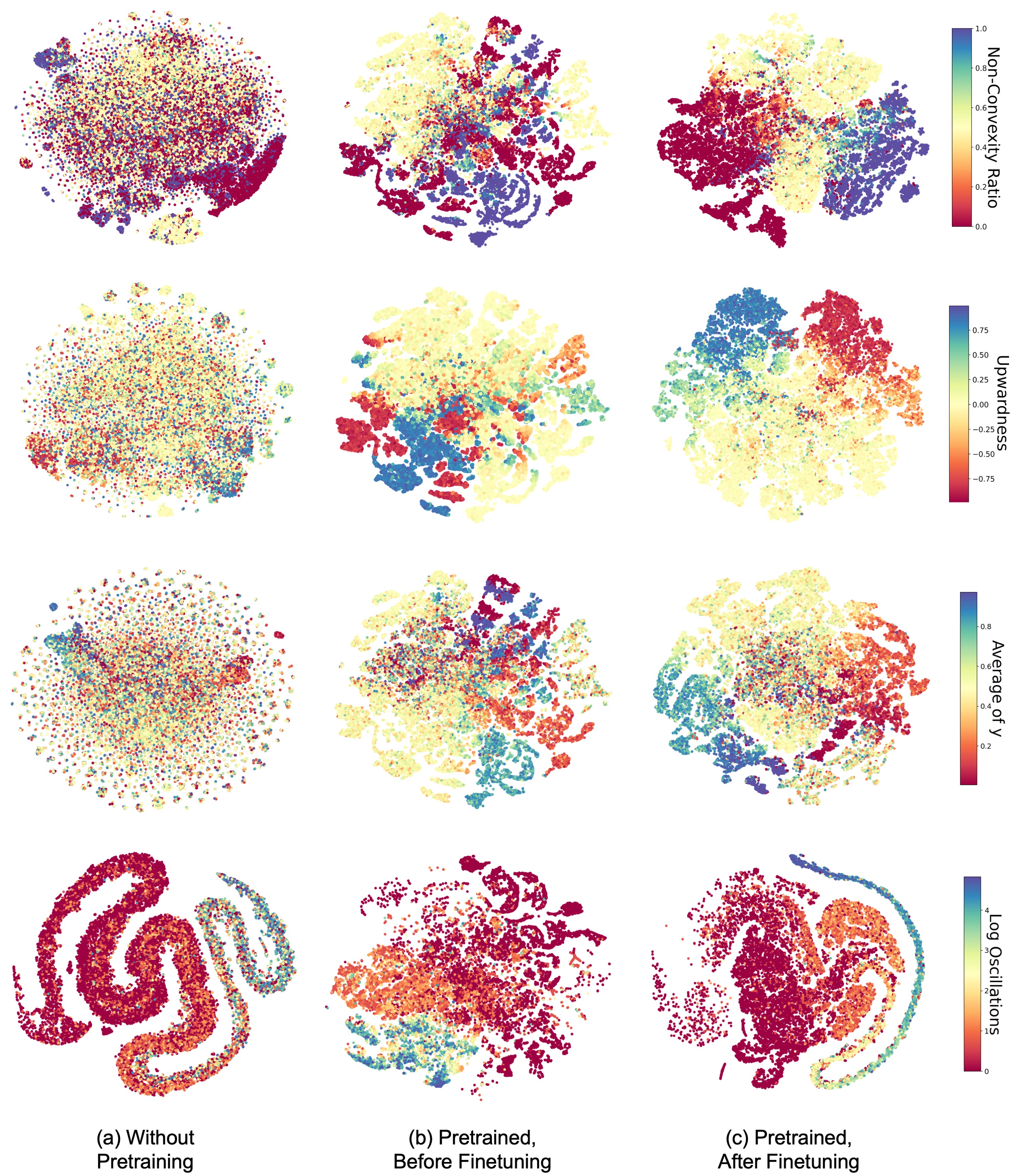}
\caption{\small 2D t-SNE plots of the symbolic encoded representations for the tasks of predicting numeric mathematical properties: Non-Convexity Ratio, Function Upwardness, Average of $y$, and Oscillations. The plots compare the \textbf{(a)} supervised models without pre-training, \textbf{(b)} frozen pre-trained \modelname encoder, and \textbf{(c)} fine-tuned \modelname encoders for each task.}
\vspace{-1.0em}
\label{fig:latentprop_appendix}
\vspace{-0.5em}
\end{figure}

In addition to numerical results, we include visual representations of the model's latent features for each property. These visualizations offer a qualitative perspective on how our model captures and represents the underlying characteristics of different properties in representations.
% Here, we analyze the properties mentioned in the previous section in the latent space. 
Fig.~\ref{fig:latentprop_appendix} shows a qualitative comparison of pre-finetuning and post-finetuning latent spaces of \modelname against that of supervised task prediction models, using 2-dimensional t-SNE visualizations of the encoded representations. The first two rows (NCR and Upwardness) are replicated from the main body (Fig.~\ref{fig:latentprop}) for ease of comparison. In each task (row), the plots are colored by the values of the corresponding property. In each task, a training dataset with $10$K samples was used to train the model. 

The observations from Fig.~\ref{fig:latentprop_appendix} show that the latent spaces of supervised models (without pre-trained \modelname) are very weakly structured and barely exhibit a recognizable trend for the properties. On the other hand, when the pre-trained \modelname is used, the latent spaces are shaped by the symbolic-numeric similarities of the functions such that numeric properties can be clustered and/or show visible trends in the symbolic encoded representation space $\bm{Z}_S$. Furthermore, fine-tuning the encoder, as shown in Fig.~\ref{fig:latentprop_appendix}(c), leads to more organized latent spaces with distinct linear property trends.

% the 2-dimensional t-SNE plots of the \modelname encoded representation (1-dimensional numeric datasets) colored by different properties. We can observe that the latent space is shaped by the symbolic-numeric similarities of the functions such that numeric properties can be clustered and/or show visible trends in the symbolic encoded representation space $Z_f$ (Figure \ref{fig:latentprop_1d_appendix}, a-d). Also the numeric encoded space $Z_y$ is organized by the symbolic properties (Figure \ref{fig:latentprop_1d_appendix}, e-f). 

% We analyze the properties mentioned in the previous section in the latent space. Figure \ref{fig:latentprop_1d_appendix} shows the 2-dimensional t-SNE plots of the \modelname encoded representation (1-dimensional numeric datasets) colored by different properties. We can observe that the latent space is shaped by the symbolic-numeric similarities of the functions such that numeric properties can be clustered and/or show visible trends in the symbolic encoded representation space $Z_f$ (Figure \ref{fig:latentprop_1d_appendix}, a-d). Also the numeric encoded space $Z_y$ is organized by the symbolic properties (Figure \ref{fig:latentprop_1d_appendix}, e-f). 

% \input{Arxiv/fig_table_tex/fig_latent_properties_1d}

%%% Comment out for compiling speed

\section{Additional Visualizations of \modelname Pre-trained Latent Space}

\begin{figure}[!ht]
\centering
\includegraphics[width=0.95\linewidth]{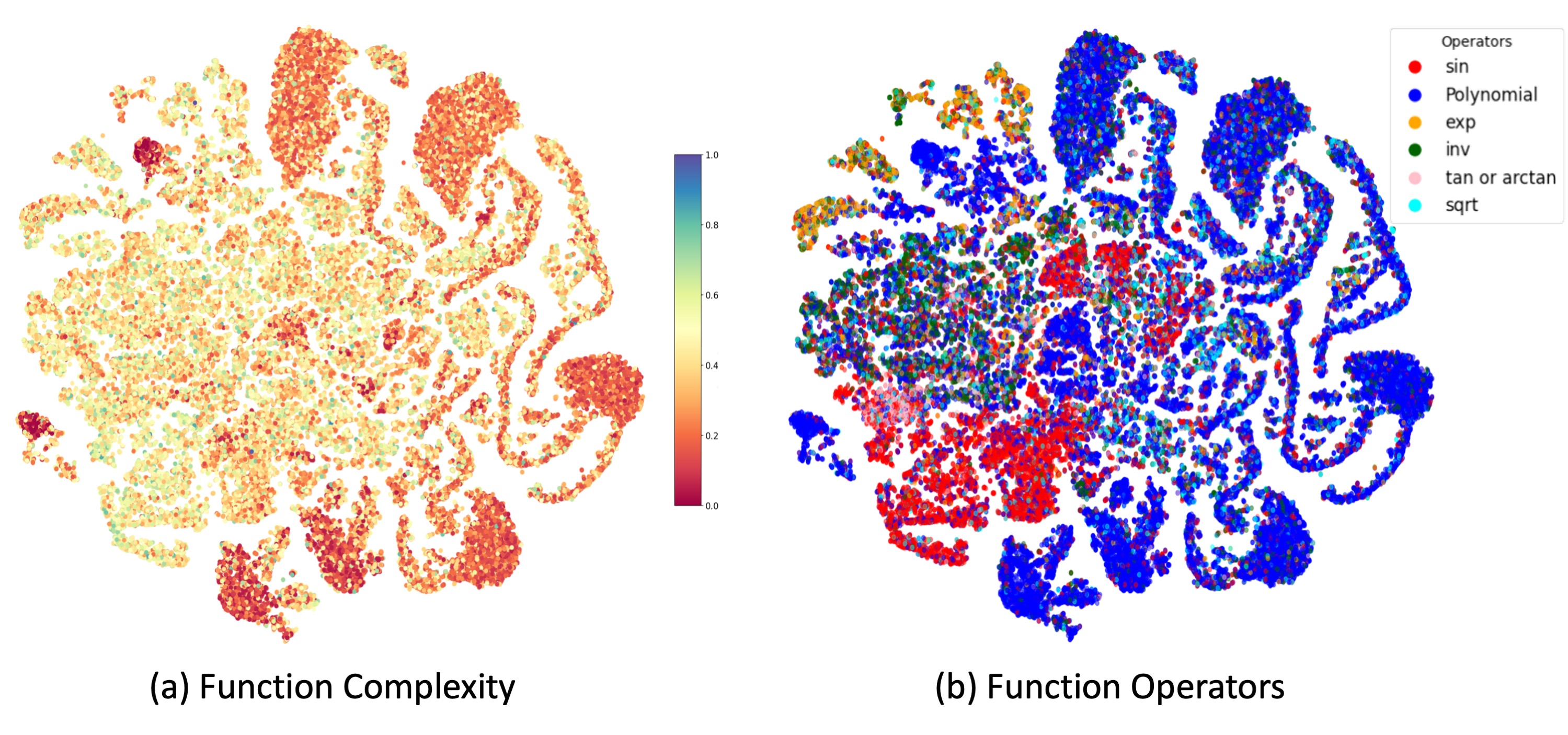}
% \captionsetup{font=footnotesize}
\caption{\small 2D t-SNE plots of the pretrained \modelname numeric encoded representations ($\bm{Z}_V$) colored by \textbf{(a)} Function Complexity, and \textbf{(b)} Function Classes based on Operators.
\vspace{-1.0em}
}
\label{fig:latentprop_class_comp}
\end{figure}

\begin{figure}[!ht]
\centering
\includegraphics[width=0.7\linewidth]{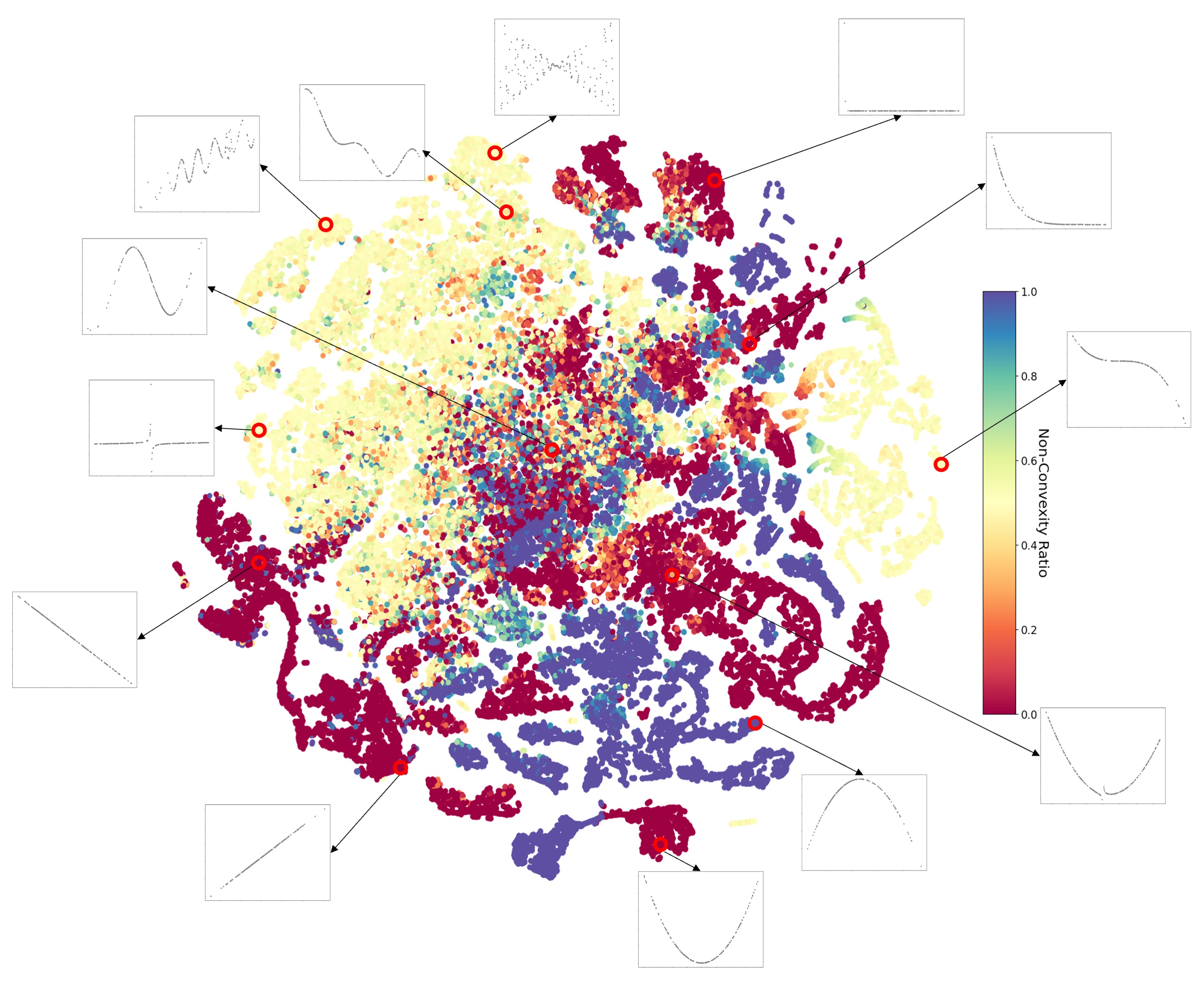}
\vspace{-0.5em}
% \captionsetup{font=footnotesize}
\caption{\small 2D t-SNE plot of the pretrained \modelname symbolic encoded representations ($\bm{Z}_S$) colored by Non-Convexity Ratio property. Adjacent to the corresponding locations of points in the latent space, the numeric behaviors of selected sample equations are displayed, illustrating the interplay between their symbolic forms and numeric properties. This visualization underscores how both the symbolic and numeric characteristics of functions influence their representation in \modelname's latent space.
% The numeric behavior of several sample equations is shown next to their relative position in the latent space. Both symbolic and numeric aspects of functions affect the latent vectors.
% \vspace{-1.0em}
}
\label{fig:latentprop_annotated}
\end{figure}

\begin{figure}[!ht]
\centering
\includegraphics[width=\linewidth]{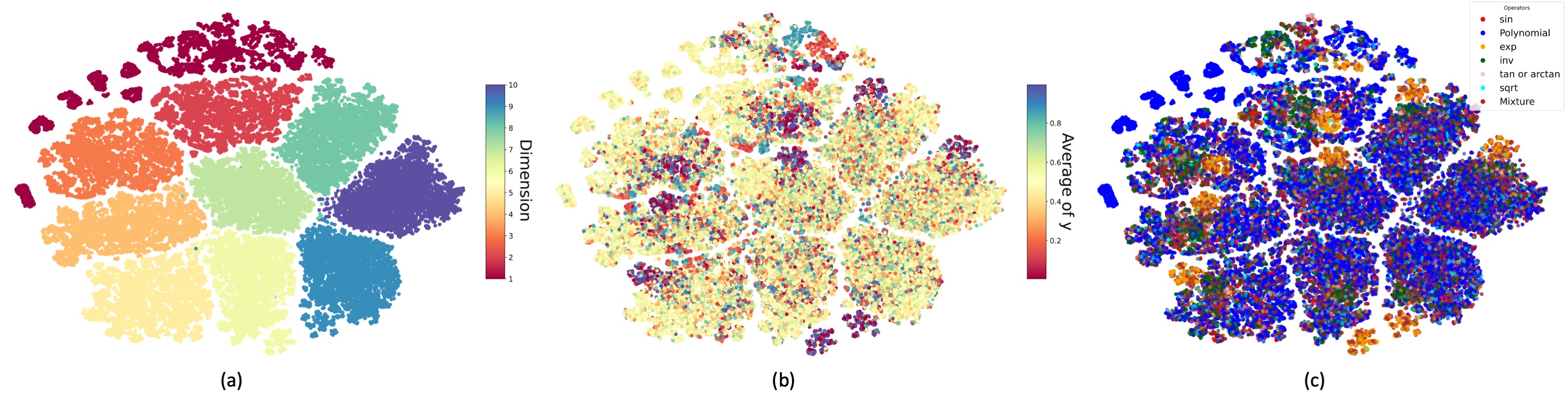}
% \captionsetup{font=footnotesize}
\caption{\small 2D t-SNE plots of the \modelname symbolic encoded representations on up to 10-dimensional datasets, colored by \textbf{(a)} dimension of the functions, \textbf{(b)} Average of normalized $y$, and \textbf{(c)} classes of functions based on their operators.
\vspace{-1.0em}
}
\label{fig:latentprop_10d_appendix}
\end{figure}

\paragraph{Numeric Encoded Representations.} 
We show that similar to how symbolic encoded representations are shaped by numeric behaviors, the numeric encoded vectors $\bm{Z}_V$ are likewise influenced by the symbolic attributes of the corresponding governing equations. Fig. \ref{fig:latentprop_class_comp} showcases 2D t-SNE visualizations of the learned latent space of \modelname's numeric encoded vectors, color-coded by function (a) complexity and (b) an arbitrarily defined categorization of the functions based on their dominant operators. Further details regarding these two symbolic features are provided below:

\textit{Function Complexity:} Function complexity, as defined in Symbolic Regression (SR) tasks, pertains to the length of the function expressed in prefix order notation,i.e., the number of nodes in the expression tree. Intuitively, functions with a greater number of operators and variables (resulting in longer equations) are considered more complex, often exhibiting correspondingly complex behaviors.

\textit{Function Operator Classes:}
Mathematical functions can be broadly classified into different classes based on the operators utilized in their expressions, which in turn influence the behavior of the data they describe. It is important to note that a single function may incorporate multiple operators, contributing to the overall complexity of the data's behavior. Additionally, certain operators within a function may hold more significance than others, exerting greater influence on the range and pattern of the data. To categorize the functions, we employ the following guidelines:

First, we consider a prioritized set of unary operators: $\mathcal{O} = \{\arctan, \tan, \exp, \rm{sqrt}, \rm{inv}, \cos, \sin, $  $\rm{pow3}, \rm{pow2} \}.$ If a function exclusively employs one of these operators, it is categorized accordingly. For simplicity, we designate both \(\rm{pow2}\) and \(\rm{pow3}\) as \texttt{Polynomial}, and we employ \texttt{sin} for both \(\sin\) and \(\cos\). In the event that a function incorporates more than one operator, it is assigned to the category corresponding to the operator of higher priority. It is worth noting that this categorization may not always perfectly capture the behavior of functions, as an operator with lower priority may potentially exert a more dominant influence than another prioritized operator.

\paragraph{Annotated Latent Space.} 
% \vspace{-0.7em}
To have a closer look to the latent space representation, we also analyze several functions with their position in the learned latent space t-SNE visualization. Fig.~\ref{fig:latentprop_annotated} shows the same t-SNE plot of $\bm{Z}_S$ (from the symbolic encoder) colored by NCR property and annotated by the numeric behavior (scaled $y$) of some samples. We can observe that the latent space is shaped by both symbolic input $f(\cdot)$ and numeric data, such that closer points have more similar symbolic and numeric features. 

% \vspace{-0.7em}
\paragraph{10-Dimensional \modelname Latent Space Analysis.} 
% \vspace{-0.7em}
Fig.~\ref{fig:latentprop_10d_appendix} shows the latent space representation of the pre-trained \modelname with numeric datasets of up to 10 dimensions, which is used for the symbolic regression task (so that we can evaluate on SRBecnh and compare with SOTA baselines). We observe that the model can cluster the functions with different dimensions, and within each cluster, it is shaped by the symbolic-numeric similarity of the functions.

%%%% Comment out for compiling speed 

\vspace{-0.5em}
\section{Details of Using \modelname for Symbolic Regression}
\label{sec:app-sr}
\vspace{-0.5em}

% \vspace{-0.5em}
% \subsection{SR Model Architecture Details}
% \vspace{-0.5em}

% \vspace{-0.5em}
\subsection{Implementation Details}
\label{sec:app-srimp}
\vspace{-0.5em}
In this section, we provide the details of the model and training procedure for the symbolic regression task. As illustrated in Fig. \ref{fig:snipsrlso} of the main body, the training step includes learning a mapping module and fine-tuning an expression generation decoder which is borrowed from \citep{Kamienny-E2E-symbolic-NIPS-2022}. We elaborate upon each of the modules and the details of training. 

\vspace{-0.3em}
\paragraph{Expression Decoder.} 
\vspace{-0.3em}
The pre-trained expression decoder from \citep{Kamienny-E2E-symbolic-NIPS-2022} is a seq2seq transformer decoder \citep{Attention-NeurIPS-2017} with 16 attention heads and the same embedding dimensionality of 512. The decoder has 16 layers (deeper compared to the encoders) to enhance its generation capacity. 

% \vspace{-0.3em}
\paragraph{Mapping Network.} 
\vspace{-0.7em}
The learnable Mapping Network $g_{\gamma}$ translates \modelname's numeric embeddings $\bm{Z}_V$ into a compatible input for the decoder $\mathcal{D}^{E2E}_{\phi}$. Therefore, we can use the power of both pre-trained encoder and decoder modules by learning a mapping between these two modules. 
In fact, $g:\mathbb{R}^{d_{\text{emb}}} \rightarrow \mathbb{R}^{L \times d_{\text{emb}}}$ reshapes \modelname embeddings into a sequence with maximum length $L$. To do so, we use a simple Multi-Layer Perceptron (MLP) design with two linear layers. The first layer applies a linear mapping from $\mathbb{R}^{d_{\text{emb}}}$ to $\mathbb{R}^{L d_{\text{emb}}}$, followed by a ReLU activation. This output is reshaped to add the sequence dimension $\mathbb{R}^{L \times d_{\text{emb}}}$ , and then passed to the second layer, which applies a linear mapping from $\mathbb{R}^{d_{\text{emb}}}$ to $\mathbb{R}^{d_{\text{emb}}}$. Consequently, the final output retains the shape $\mathbb{R}^{L \times d_{\text{emb}}}$.

% \vspace{-0.3em}
\paragraph{Training} 
\vspace{-0.7em}
Similar to the suggestions of \citep{mokady2021clipcap}, we found that to effectively learn the simple MLP mapping network, we can let the decoder network to be simultaneously fine-tuned. In this way, the mapping training is less challenging since we have a control over both networks. We train the whole model in two stages. In the first stage, we freeze the \modelname encoder's parameters and only update the mapping network and decoder's parameters. This allows the model to learn the mapping from the fixed encoded representations of numeric datasets to their corresponding symbolic functions. Similar to the pre-training procedure, an Adam optimizer with learning rate warm-up followed by a inverse square root decay based on number of steps is used to train the model with cross-entropy loss. In the second stage, to enhance the model's generation capacity, we fine-tune the \modelname's encoder along with the other modules. This helps the model to distinguish between the overlapped representations in the encoder, which were not originally trained for the expression generation objective. It also maintains their relative positions obtained from the contrastive loss. In both stages, we use batch size $B=128$ for training.
% from $10^{-7}$ to $4.10^{-5}$ for the first 10,000 steps

\vspace{-0.5em}
\subsection{\modelname Latent Space Optimization Details}
\label{sec:app-srlso}
\vspace{-0.5em}
In this section, we provide the details of the Latent Space Optimization (LSO) on \modelname's encoded representations. This method combines three main advantages that make it suitable for the symbolic regression task. 
\vspace{-0.5em}

\begin{itemize}
    \item By training an expression decoder on top of \modelname encoder, we learn a prior for function generation given the numeric dataset, which is the main advantage of neural symbolic regression models over traditional search methods.
    \vspace{-0.4em}

    \item While neural SR models are trained using token-matching objectives, LSO utilizes a powerful search with the objective of fitting accuracy. Therefore, it can also enjoy the main advantage of the search methods over the pre-trained equation generation methods. 
    \vspace{-0.4em}

    \item The most important advantage of this method is that it \textbf{exploits the well-organized latent space of \modelname to perform the optimization in a continuous, low-dimensional, and interpolatable latent space} which provides it with a huge benefit over traditional GP functions search techniques.
\end{itemize}

\vspace{-0.5em}

Algorithm \ref{alg:snip_lso} sketches the main steps of LSO. The \textcolor{purple}{red} lines indicate when the modules of pre-trained model are called, and \textcolor{blue}{blue} lines indicate when other functions are called. 

{
\centering
\begin{minipage}{0.9\linewidth}
\begin{algorithm}[H]
\fontsize{9}{9}\selectfont
\SetAlgoLined
\SetKwInOut{Input}{Input}
\caption{Latent Space Optimization (LSO) on SNIP Pre-trained Encodings}
\label{alg:snip_lso}
\Input{Dataset $\{\bm{x}, \bm{y} \}$, sampling size $b$, stopping $R^2_{stop}$, Maximum Iterations $T$}
\BlankLine
\textcolor{brown}{1. Population Generation}\\
\textbf{Generate the search population by following the steps below:}
\begin{itemize}
    \item Generating $p_1<P$ points by randomly sampling subsets of the original dataset and \textcolor{purple}{Encoding} $\bm{Z}_{V}^{i}$ for $i \in \{1, \dots, p_1\}$.
    \item Generating $p_2<P$ points by injecting Gaussian noise to the input data points and \textcolor{purple}{Encoding} $\bm{Z}_{V}^{i}$ for $i \in \{1, \dots, p_2\}$.
    \item Generating $p_3<P$ points by first \textcolor{purple}{Encoding} a fixed input dataset and then injecting Gaussian noise to the encoded representation $\bm{Z}_V$ to get $\bm{Z}_{V}^{i}$ for $i \in \{1, \dots, p_3\}$.
    \item Combining these points to have a population with size $P = p_1 + p_2 + p_3$.
\end{itemize}
\For {$T$ iterations}{
    \textcolor{brown}{2. Fitness Evaluation}\\
    \For {each population agent $\bm{Z}_{V}^{i}$, $i \in \{1, \dots, P\}$}{
        Compute the fitness value following steps below:\\
        1. \textcolor{purple}{Decode} the encoded representation with sampling size $b$ to get $\{\tilde{f}_1^{i}, ..., \tilde{f}_b^{i}\}$.\\
        \vspace{0.1em}
        2. Remove functions with duplicate skeletons.\\
        3. \textcolor{blue}{Refine} the constants of the remaining functions on the training set using BFGS.\\
        4. Compute $R^2$ score for each function; Store the highest score as the fitness value of the agent at current iteration $\mathcal{F}^{i}_{t}$.\
    }
    Set $\mathcal{F}^{*}_{t} = \max_i (\mathcal{F}^{i}_{t})$ as the best current score. \\
    \If {$\mathcal{F}^{*}_{t} > R^2_\text{stop}$}{
        Return the best function $f^*(\cdot)$ to be evaluated on the testing dataset.\\
    }
    \textcolor{brown}{3. Optimization Step}\\
    Call the \textcolor{blue}{Gradient-Free Optimizer} update rule with $(\bm{Z}_V , \mathcal{F}_{t} )$ to get the updated population $\bm{Z}_V$.\\
}
Report the best function $f^*(\cdot)$, and evaluate it on the testing dataset.
\end{algorithm}
\end{minipage}
\par
}

Some of the details of these steps are as follows:
\vspace{-0.5em}
\paragraph{Population Generation.} To combine the use of prior knowledge with the search method, instead of generating random agents in the latent space, we initialize the population by augmenting the given dataset. In algorithm \ref{alg:snip_lso}, $p_1$, $p_2$, and $p_3$ are selected to be 15, 10, and 25, respectively, summing up to $P=50$ to maintain a balance on the performance and the computation time. Each of the augmentations provides a different perspective that we elaborate upon:
\vspace{-0.5em}
\begin{itemize}
    \item In the first augmentation, $\mathcal{P}_1$, each augmented agent $\bm{Z}_V^{i}$ is obtained by first uniformly sampling a subset, with size $n < N$ of the original dataset $\left(\bm{x}^{\text{sub}_{i}},\bm{y}^{\text{sub}_{i}}\right) \subseteq \left(\bm{x},\bm{y}\right) $. Since the maximum sequence length is 200, we set $n=200$ if $N>400$, and set $n = \lfloor N/2 \rfloor$ if $N<400$. Subsequently, we encode the sampled data to get $\bm{Z}_V^{i} = \mathcal{E}^V_{\theta} \left( \{\left(\bm{x}^{\text{sub}_i},\bm{y}^{\text{sub}_i} \right)\} \right)$.
\vspace{-0.4em}
    \item For the second augmentation, $\mathcal{P}_2$, each augmented agent $Z_V^{i}$ is obtained by first perturbing the target values with random Gaussian noise $\left(\bm{x},\bm{y}+\bm{\epsilon}_i \right) $, where $\bm{\epsilon}_i \sim \mathcal{N}(0, \sigma_i^2 I_{n})$, and $\sigma_i \propto i$ to cover different ranges of perturbations for a more diverse search population. Subsequently, we encode the perturbed data to get $\bm{Z}_V^{i} = \mathcal{E}^V_{\theta} \left( \{\left(\bm{x},\bm{y}+\epsilon_i \right)\} \right)$.
\vspace{-0.4em}
    \item For the third augmentation, $\mathcal{P}_3$, each augmented agent $Z_V^{i}$ is obtained by first encoding the dataset to get $\bm{Z}_V = \mathcal{E}^V_{\theta} \left( \{\left(\bm{x},\bm{y}\right)\} \right)$, and then perturbing the encoded vectors using random Gaussian noise. $ \bm{Z}_V^{i} = \bm{Z}_V + \mathcal{N}(0, \sigma_i^2 I_{d_{\text{emb}}})$, where $\sigma_i$ varies randomly to achieve a more diverse search population.
\end{itemize}

\paragraph{Fitness Evaluation.} To evaluate the fitness of the population $\mathcal{P}$ at iteration $t$, we utilize the expression generation modules with sampling \citep{sampling-2018-hierarchical} to generate $b=2$ candidates for each agent $\bm{Z}_V^{i}$. Following this, candidates with duplicate skeletons are eliminated, and the remaining candidate skeletons undergo a refinement process. In order to refine the constant values, a procedure following \citep{Kamienny-E2E-symbolic-NIPS-2022} is employed. Specifically, the generated constants (model predictions) serve as initial points, and these constants are further optimized using the BFGS algorithm \citep{bfgs_flet87}. Subsequently, we calculate the $R^2$ score on the training data points, which serves as the fitness values for the population.

% To evaluate the fitness of the population $\mathcal{P}$ at iteration $t$, we call the expression generation modules with sampling \citep{sampling-2018-hierarchical} to generate $b$ candidates for each agent $\bm{Z}_V^{i}$. The candidates with duplicate skeletons are then removed, and the new candidate skeletons go through a refinement process. To refine the constant values, following \citep{Kamienny-E2E-symbolic-NIPS-2022}, the generated constants (model predictions) are used as initialized points and then these constants are fine-tuned using BFGS \citep{bfgs_flet87} algorithm. Finally, we compute $R^2$ on the training data points as the population fitness values which we aim to maximize.

\paragraph{Optimization.}
Computing the fitness measure \(R^2\) from the generated equation \(\tilde{f}_i\) is not a differentiable process. Consequently, we resort to utilizing gradient-free optimization algorithms, which operate without the need for gradient information to update the search population. In this context, swarm intelligence algorithms have proven to be both computationally efficient and effective for continuous spaces. Therefore, we opt for a recently developed swarm algorithm known as the Grey Wolf Optimizer (GWO) \citep{gwo-2014} for updating the population vectors. The GWO algorithm employs a balanced exploration-exploitation strategy based on the current elite population agents, i.e., those agents exhibiting the best fitness values. In this work, we select the maximum iteration $T=80$, and we use early stopping criterion $R^2_\text{stop} = 0.99$. Also, at each iteration, we establish lower and upper bounds for agent positions based on the minimum and maximum values of $Z_V$ across both dimensions and all agents.

\subsection{SRBench Evaluation Dataset Details}
\label{sec:app-srdatasets}
% \vspace{-0.5em}

In our evaluation of \modelname, we resort to the widely-recognized SRBench, a benchmark known for its challenging and diverse datasets in Symbolic Regression \citep{SRBench-Cava-NeurIPS-2021}. This benchmark aggregates datasets from three primary groups: \textit{Feynman}, \textit{Strogatz}, and \textit{Black-box regression}. A visual representation of these datasets is presented in Fig.~\ref{fig:app-srdatasets}, illustrating the distribution across groups in terms of dataset count, input dimensions, and the number of datapoints. More details on each of these data groups are given below.

%%%%%%%%%%%%
% We evaluate \modelname and several baseline methods on the following four standard benchmark datasets: \textit{Feynman}, \textit{Black-box}, and \textit{Strogatz} from SRBench \cite{SRBench-Cava-NeurIPS-2021}. More details on each of these datasets are given below.
%%%%%%%%%%%%
% Each Penn Machine Learning Benchmark (PMLB) dataset is split into train and test sets using a $75/25$ ratio. Details on each dataset can be found below.

\vspace{-0.5em}
\begin{myitemize2}
\item 
\textbf{\textit{Feynman\footnote{\url{https://space.mit.edu/home/tegmark/aifeynman.html}}}}: The Feynman dataset is a significant component of the broader landscape of symbolic regression datasets, with its roots traced back to the renowned \textit{Feynman Lectures on Physics database} series \citep{AI-Feynman-Science-2020}.
% \noindent \underline{\textit{Equations and Scope:}} 
The dataset aggregates a collection of $119$ distinct equations, as visualized in Fig.~\ref{fig:app-srdatasets}(a). These equations encapsulate a wide range of physical phenomena.
% , serving as a testament to Feynman's contributions to the realm of physics.
% \noindent \underline{\textit{Data Points and Benchmarking:}}
The regression input points $(x,y)$ for these equations are indexed within the Penn Machine Learning Benchmark (PMLB) \citep{SRBench-Cava-NeurIPS-2021, Olson2017PMLB}. The SRBench has studied these equations, adopting them as standards in the evaluation of symbolic regression methodologies.
% \noindent \underline{\textit{Dimensions and Underlying Functions:}}
One of the critical constraints of this dataset is the input dimensionality, which has been capped at $D\leq 10$, as depicted in Fig.~\ref{fig:app-srdatasets}(b). This limit ensures a consistent evaluation scale across multiple symbolic regression challenges. Moreover, an advantage that researchers have with this dataset is the availability of the true underlying functions, eliminating the ambiguity often present in black-box datasets.
% \noindent \underline{\textit{Data Size and Bagging}}
Each dataset in this data group includes $10^5$ datapoints, as highlighted in Fig.~\ref{fig:app-srdatasets}(c).

\begin{figure}[t]
\centering
\includegraphics[width=0.9\linewidth]{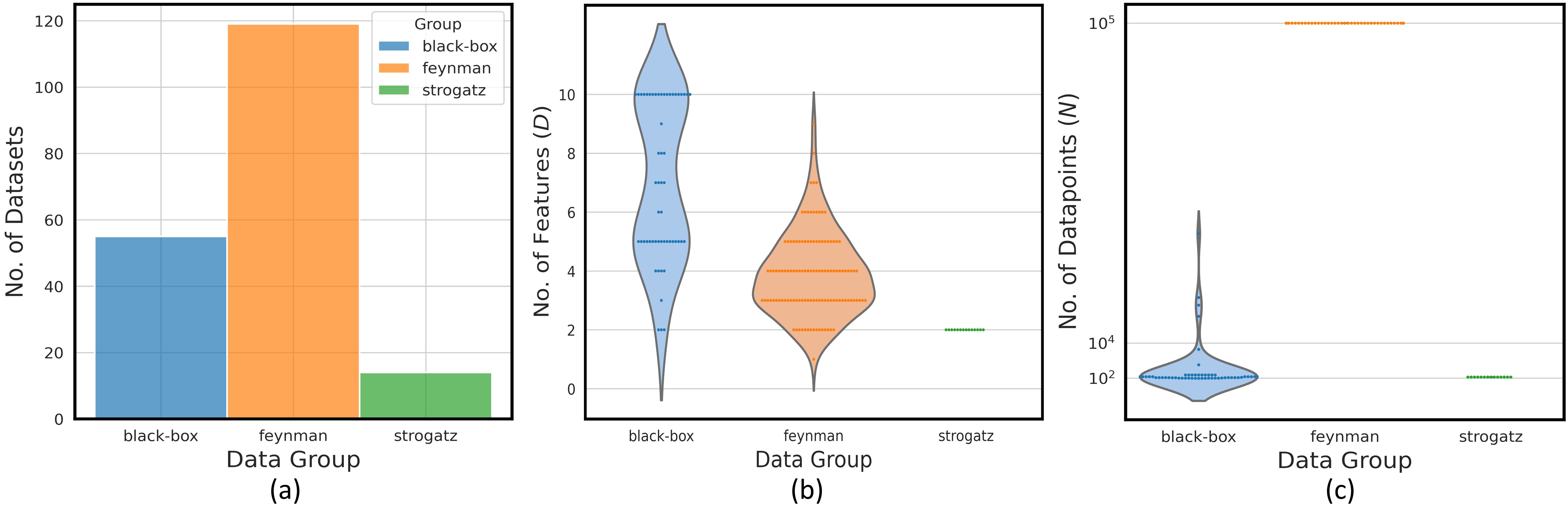}
% \captionsetup{font=footnotesize}
\caption{ Distribution of datasets across the SRBench \textit{Feynman}, \textit{Strogatz}, and \textit{Black-box} groups: (a) Count of datasets, (b) Spread of input dimensions, and (c) Number of datapoints per dataset.
% \vspace{-1.0em}
}
\label{fig:app-srdatasets}
% \vspace{-0.5em}
\end{figure}

\item \textbf{\textit{Strogatz\footnote{\url{https://github.com/lacava/ode-strogatz}}}:} 
The Strogatz dataset is a collection of symbolic regression challenges, drawing inspiration from the domain of nonlinear dynamical systems. Its inclusion in the broader context of symbolic regression evaluations offers a unique perspective, focusing on the intricacies of dynamical behaviors.
% \noindent \underline{\textit{Origin and Content:}} 
At the heart of this dataset are $14$ distinctive symbolic regression problems, extracted from the \textit{ODE-Strogatz database} \citep{LACAVA_strogatz2016}. Each of these problems has been crafted to reflect the nuances of nonlinear dynamical systems, underscoring the rich tapestry of dynamical phenomena.
% \noindent \underline{\textit{Benchmarking and Data Points:}} 
The regression input points $(x,y)$ for these challenges are accessible from the Penn Machine Learning Benchmark (PMLB) \citep{Olson2017PMLB}. The SRBench has also leveraged these problems, incorporating them into an evaluation framework for symbolic regression \citep{SRBench-Cava-NeurIPS-2021}.
% \noindent \underline{\textit{Dimensionality and Function Knowledge:}} 
An inherent attribute of this dataset is the limitation on input dimensionality, fixed at $D=2$ (as shown in Fig.~\ref{fig:app-srdatasets}(b)). This means that for each problem, there are two primary input variables. This restriction facilitates a concentrated exploration of two-dimensional dynamical systems. The true functions, which underlie and generate the data, are also available.
% \noindent \underline{\textit{Data Size and Bagging Strategy:}} 
Each problem in the Strogatz data group contains $N=400$ data points.

%%% no_bagging 
% To maintain consistency with other evaluations and due to constraints from established methodologies \citep{Kamienny-E2E-symbolic-NIPS-2022}, we adopt a similar bagging approach for the Strogatz problems.
% Each dataset undergoes partitioning into $B$ bags, with each bag designed to contain a maximum of $200$ input points.

%%%%%%%%%%%%%%%%%%%%%%
% This dataset comprises $14$ symbolic regression problems sourced from the \textit{ODE-Strogatz database} \cite{LACAVA_strogatz2016}, focusing on nonlinear dynamical systems. The input points $(x,y)$ for these problems are included in PMLB \cite{Olson2017PMLB} and have been examined in SRBench \cite{SRBench-Cava-NeurIPS-2021} for symbolic regression. The input dimension for these problems is restricted to $D=2$ and the true underlying functions are provided. Also, the dataset of each equation in this group contains $N=400$ data points. Similar to Feynman experiments, for evaluations on these datasets, we partition each dataset into $B$ bags, each accommodating maximum $200$ input points (as outlined in \citep{Kamienny-E2E-symbolic-NIPS-2022}).
%%%%%%%%%%%%%%%%%%%%%%

\item \textbf{\textit{Black-box\footnote{\url{https://github.com/EpistasisLab/pmlb/tree/master/datasets}}}:} The Black-box dataset group stands as a testament to the versatility and applicability of symbolic regression methods to real-world complex datasets without known underlying functions, offering challenges that are both diverse in nature and crucial for machine learning evaluations.
% \noindent \underline{\textit{Source and Objective:}} 
Primarily sourced from the comprehensive PMLB \citep{Olson2017PMLB}, the Black-box datasets have gained significant attention in the SRBench \citep{SRBench-Cava-NeurIPS-2021}, serving as key benchmarks against various state-of-the-art ML regression methods. The primary objective of utilizing the Black-box datasets for symbolic regression is not just about achieving fitting accuracy, but also deciphering models that are interpretable yet better-fitting (compared to ML models), offering insight into the underlying data processes.
% \noindent \underline{\textit{Constraints and Applicability:}} 
Ensuring compatibility with the latest methods, we constrain datasets to only possess continuous features as well as input dimension that doesn't exceed 10: $D\leq10$. This decision aligns with the training preconditions of the Transformer \modelname numeric encoder and \text{E2E's} SR decoder \citep{Kamienny-E2E-symbolic-NIPS-2022}, which are tailored for an upper limit of $d_{max}=10$. As a result, out of a broader set, $57$ black-box datasets meet the constraint.
% \noindent \underline{\textit{Dataset Nature and Noise Levels:}} 
Datasets of this group offer various challenges, stemming from both the real-world and synthetically generated scenarios. An inherent characteristic of these datasets is the noise, varying in levels, which mimics real-world data inconsistencies, enhancing the robustness of evaluations.
% \noindent \underline{\textit{Dataset Size and Distribution:}} 
The Black-box collection contains an impressive diversity in terms of number of data points per dataset, ranging from as low as $47$ to around $40$K. To visually represent this, Fig.~\ref{fig:app-srdatasets}(c) demonstrates the distribution of datasets across number of datapoints ($N$), with an observable average data point scale around $10^2$ for this group.
\end{myitemize2}

\begin{figure}[!ht]
\centering
\includegraphics[width=0.8\linewidth]{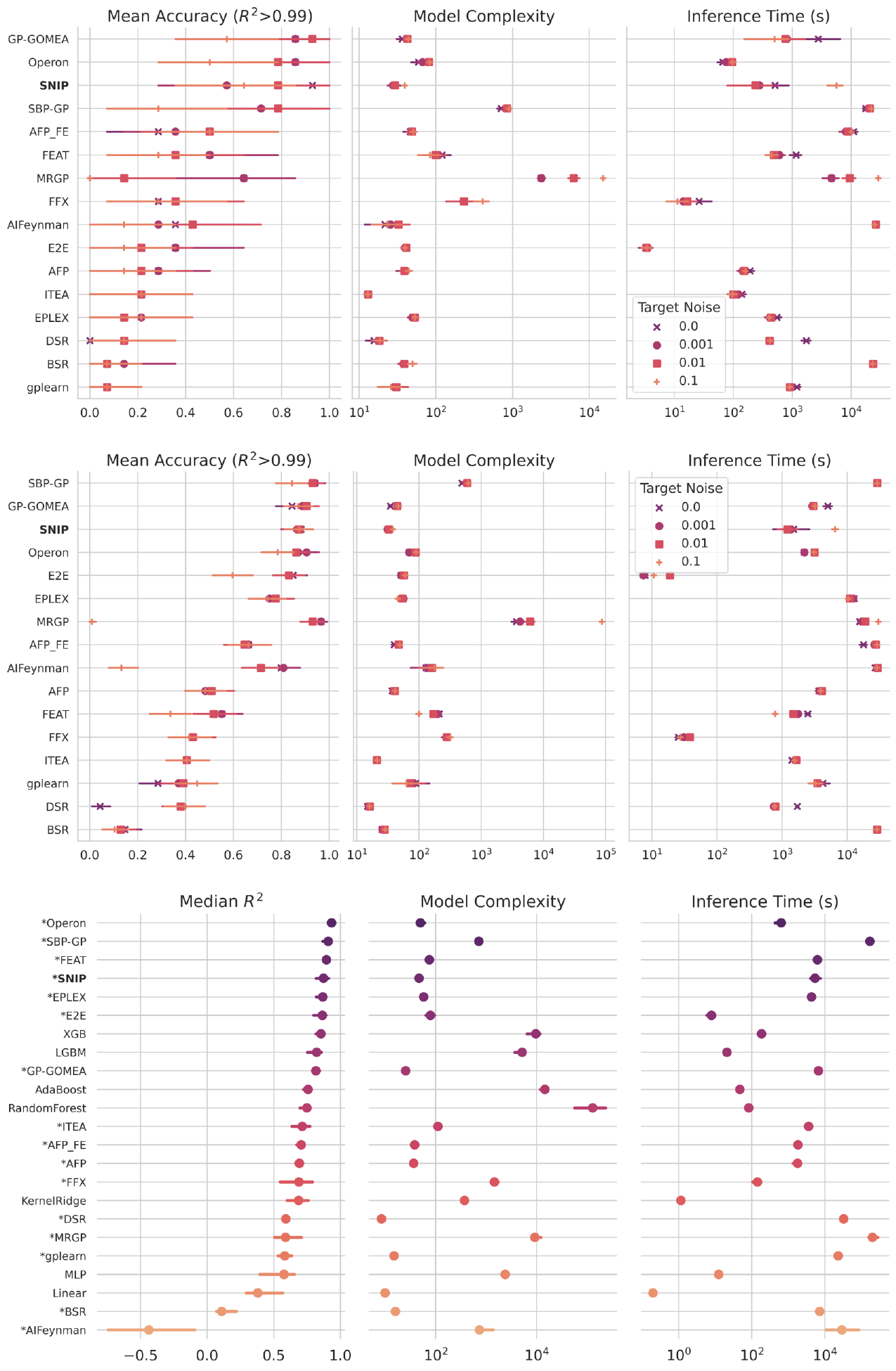}
% \captionsetup{font=footnotesize}
\caption{Performance comparison of \modelname and SRBench algorithms in terms of Accuracy-Complexity-Time on \textbf{\textit{Strogatz}} (\textbf{top}), \textbf{\textit{Feynman}} (\textbf{middle}), and \textbf{\textit{Black-box}} (\textbf{bottom}) datasets. For \textit{Feynman} and \textit{Strogatz} dataset, algorithms are sorted based on mean accuracy defined as the ratio of solutions with $R^2>0.99$ on test set under various noise levels, and for \textit{Black-box} datasets, the algorithms are sorted based on the median $R^2$ score on test set. \modelname demonstrates a strong balance of performance with relatively low model complexity and competitive inference time compared to GP-based algorithms. The error bars represent the 95\% confidence interval and "$*$" refers to SR methods for \textit{Black-box} dataset.
\vspace{-1.0em}
}
% \vspace{-0.5em}
\label{fig:srpairgrid}
\end{figure}

% \vspace{-0.5em}
\subsection{Additional Details for SRBench Evaluation Experiments}
\label{sec:app-srexps}

\subsubsection{Experiment Settings}
\label{sec:app-srsettings}

Aligning with the SRBench evaluations and the methodology from \citep{Kamienny-E2E-symbolic-NIPS-2022}, we partition the observation points of each equation in the SRBench datasets (comprising \textit{Feynman}, \textit{Strogatz}, and \textit{Black-box}) into training and testing subsets with a $75\%/25\%$ split.

\subsubsection{Results}
\label{sec:app-srresults}
\vspace{-0.5em}
\paragraph{Strogatz:}
In Fig. \ref{fig:srpairgrid} (\textbf{top}), we compare the performance of \modelname with the SRBench algorithms on the \textit{Strogatz} dataset. As described in Sec.~\ref{sec:app-srdatasets}, the \textit{Strogatz} dataset includes 14 equations from a two-state system governed by a first-order ODE. A key observation is that the end-to-end (E2E) transformer SR model underperforms on this dataset compared to other GP-based models. This underperformance can be attributed to the distinct time-ordered distribution of observations in the \textit{Strogatz} dataset, which deviates considerably from the E2E model's pre-training data. Interestingly, \modelname, despite not being trained on time-ordered data, significantly outperforms not only the E2E transformer SR model but also many other leading SR baselines. In terms of accuracy, \modelname ranks within the top three baselines, specifically when evaluating the proportion of solutions with an $R^2~>~0.99$ across varying target noise levels. Its inference time is competitive with leading baselines such as GP-GOMEA and Operon. \modelname also obtains a very good model complexity. It produces expressions that fit well but with lower complexity than top-ranked competitors in Fig.\ref{fig:srpairgrid}.

\begin{wrapfigure}[36]{r}{0.55\linewidth}
% \begin{figure}[t]
\centering
\vspace{-1.0em}
\includegraphics[width=0.95\linewidth]{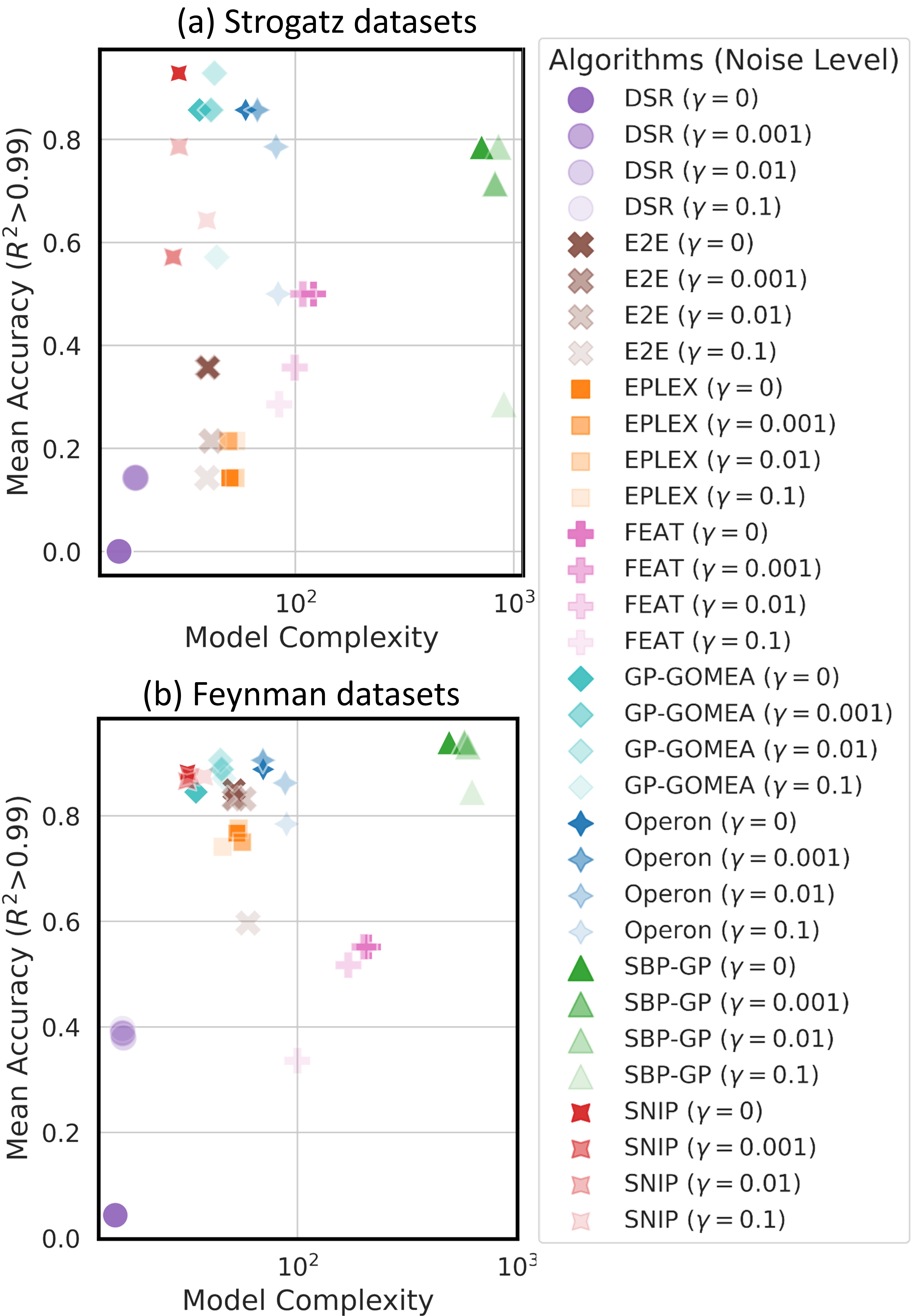}
% \captionsetup{font=footnotesize}
\caption{Pareto analysis on \textbf{(a) \textit{Strogatz}} and \textbf{(b) \textit{Feynman}} datasets, comparing methods based on fitting accuracy (proportion of solutions with $R^2>0.99$) and equation complexity across different noise levels. In both datasets, \modelname mostly stays in the desired upper-left corner, showcasing its robustness in balancing fitting accuracy and complexity even when noise is introduced. 
% Pareto comparison of competing methods for fitting-complexity trade-off with different noise levels across \textbf{(a) \texit{Feynman},} and \textbf{(b) \textit{Strogatz} datasets}. Here, the y-axis shows the percentage of good-fitting equations, i.e., $Mean(R^2>0.99)$, and the x-axis shows the mean of equation complexity. Better fitting and complexity is equivalent to greater $Mean(R^2>0.99)$ and lower eqution complexity. So, the desired place for a good equation is the upper-left position. As we can see, on both data groups, \modelname is placed on the upper-left position when there's no noise, i.e., $\gamma=0$, showing a strong fitting-complexity trade-off. As noise level increases, the performance of all methods degrades and we can see that \modelname still has a competitive performance in case of various noise levels.  Among all competing methods, it seems that \texttt{E2E} is less robust/ most vulnerable to injected noise, probably because it only relies on the priors learned with large-scale training and lacks the search process, which other methods do, for each new dataset.  
\vspace{-1.0em}
}
\vspace{-0.5em}
\label{fig:srnoisepareto}
% \end{figure}
\end{wrapfigure}
This advantage in balancing accuracy and complexity is also evident in 
Fig.\ref{fig:srbenchpareto}(a), where \modelname is positioned on the first Pareto-front, while competitors like GP-GOMEA and Operon fall on the second and third, respectively. This suggests that \modelname offers a superior Accuracy-Complexity trade-off for noise-free data~($\gamma=0$). 
Fig.~\ref{fig:srnoisepareto}(a) further underscores this point by illustrating the Pareto performance of leading SR baselines on the \textit{Strogatz} dataset across various noise levels. As expected, all methods experience a performance drop as target noise increases. Yet, \modelname consistently maintains its advantageous position in the upper-left corner, indicating its ability to generate better expression both in accuracy and complexity.

\vspace{-1.0em}
\paragraph{Feynman:}
In Fig. \ref{fig:srpairgrid} (\textbf{middle}), we present a comparative analysis of \modelname against the SRBench algorithms on the \textit{Feynman} dataset. 
As outlined in Sec.~\ref{sec:app-srdatasets}, the \textit{Feynman} dataset encompasses 119 unique Feynman equations, representing a broad spectrum of physical phenomena. This figure delineates the positioning of each algorithm in terms of Accuracy-Complexity-Time. Notably, the E2E transformer model exhibits enhanced performance on the \textit{Feynman} dataset relative to the Strogatz dataset, achieving a fifth rank in accuracy. 
\modelname, however, surpasses the performance of not only the E2E transformer SR model but also many top GP baselines. When focusing on accuracy, \modelname locates among the top three baselines, especially when considering solutions with an $R^2~>~0.99$ across diverse target noise levels. \modelname often outperforms other competing baselines like GP-GOMEA, SBP-GP, and Operon in terms of inference time. From a complexity perspective, \modelname exhibits better results against SBP-GP and Operon and demonstrates comparable performance with the GP-GOMEA baseline. This optimal balance between accuracy and complexity was also shown in Fig.\ref{fig:srbenchpareto}(c), showcasing ranking Pareto plots for both metrics. 
Here, \modelname positions on the first Pareto-front — an indicator of its better ranking in both accuracy and complexity. 
Among competitors, GP-GOMEA is positioned on the secondary Pareto level, denoting an inferior Accuracy-Complexity balance relative to \modelname. While Operon and SBP-GP are also placed on the first Pareto-dominance, their placement drifts further 
% from the coveted lower-left corner 
to the upper-left corner than \modelname — suggesting the expressions they generate reaches higher complexity. This placement shows \text{\modelname's} strong Accuracy-Complexity balance for data without noise ($\gamma=0$). To accentuate this advantage, Fig.~\ref{fig:srnoisepareto}(b) offers a deep dive into the Pareto Accuracy-Complexity performances of leading SR baselines on the \textit{Feynman} dataset across a spectrum of target noise levels. Predictably, increased noise compromises the performance of all algorithms. Still, \modelname consistently holds a favorable position, indicating it generates more accurate and less complex expressions even with increased noise.

\vspace{-1.0em}
\paragraph{Black-box:}
The study by SRBench \citep{SRBench-Cava-NeurIPS-2021} delved into black-box problems, which were initially derived from OpenML\footnote{\url{https://www.openml.org/}} and later incorporated into the PMLB datasets \citep{Olson2017PMLB}. The intent behind assessing SR models on this dataset revolves around understanding how SR methods measure up against conventional well-known machine learning techniques, especially when faced with real-world, potentially noisy or sparse datasets. As delineated in Sec.~\ref{sec:app-srdatasets}, the \textit{Black-box} dataset used here comprises 57 regression datasets. In Fig. \ref{fig:srpairgrid}~(\textbf{bottom}), we compare \modelname to the SRBench algorithms using this dataset. The figure shows how each algorithm performs in terms of accuracy, complexity, and inference time. 
% Notably, the prefix ``*'' preceding algorithm names serves to differentiate SR methods from ML methods.
The “*” sign before some method names means they're SR methods; others are machine learning methods. From the figure, it's clear that \modelname does better than both the E2E transformer SR model as well as other leading SR methods. For accuracy on Black-box datasets, measured by the median $R^2$ score in SRBench, \modelname is ranked fourth among all methods. Its inference time on the Black-box dataset is similar to most other competing methods. Diving into complexity, \modelname outperforms many of the top-tier baselines. Specifically, \modelname provides an average complexity score of $47.52$ on the Black-box datasets, which is better than its counterparts like Operon ($64.95$), SBP-GP ($639.19$), FEAT ($74.18$), EPLEX ($55.82$), and E2E ($82.78$). Further analysis of this balanced accuracy and complexity is also provided in Fig.\ref{fig:srbenchpareto}(b), which presents Pareto plots capturing both dimensions for the Black-box datasets. Consistent with earlier observations, \modelname holds its position on the first Pareto-front for this dataset. On the other hand, several competitors, including SBP-GP, FEAT, and EPLEX, are located on the secondary Pareto level, with E2E placing even further on the third. Such placements underscore their relative shortcomings in balancing accuracy with complexity compared to \modelname. While GP-GOMEA shares the first Pareto-front with \modelname, its fitting accuracy is worse, ranking 9th in Fig.~\ref{fig:srpairgrid}–falling behind conventional ML methods like XGBoost and LGBM.  So, out of all methods, Operon is the closest competitor to \modelname on these datasets. Operon fits Black-box datasets slightly better with a score of $0.933$ compared to \text{\modelname’s} $0.872$, but it offers more complex expressions with a score of $64.95$ against \text{\modelname’s} simpler $47.52$.

\subsection{Additional Results on the In-domain Synthetic Datasets.}
\label{sec:app-srexps2}

% \newpage

% \begin{figure}[t]
% \centering
% % \vspace{-1.5em}
% \includegraphics[width=\linewidth]{Arxiv/fig_table_file/r2_detailed_indomain.jpg}
% \vspace{-1.0em}
% \caption{\textcolor{blue}{\small 
% Detailed Performance Evaluation of \textbf{Mean Accuracy} between \modelname and E2E Transformer across 400 Synthetic Validation Functions. Mean accuracy is assessed by the proportion of solutions with an $R^2$ score greater than $0.99$, across different levels of formula and input difficulties: \textbf{(a) number of binary operators, (b) number of unary operators, (c) input dimension, and (d) number of input centroids.} 
% Performance comparison of \modelname and E2E transformer baseline in terms of mean accuracy defined as the ratio of solutions with $R^2>0.99$ over $400$ in-domain synthetic validation functions
% across different levels of formula and input difficulties: \textbf{(a) number of binary operators, (b) number of unary operators, (c) input dimension, and (d) number of input centroids.}
% }}
% \label{fig:sr-indomain-fit}
% \vspace{-0.5em}
% \end{figure}

\begin{figure}[t]
\centering
% \vspace{-1.5em}
\includegraphics[width=\linewidth]
{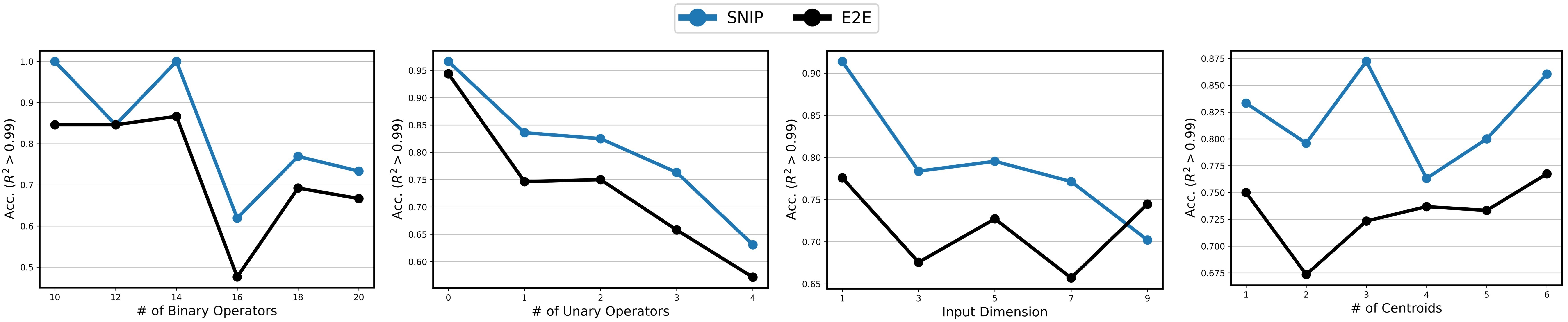}
% \vspace{-1.0em}
\includegraphics[width=\linewidth]
{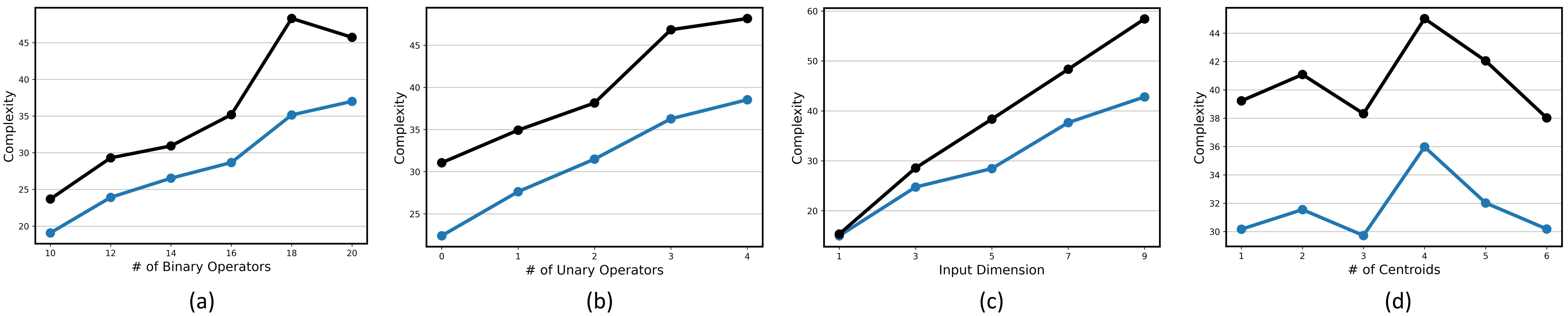}
\vspace{-1.0em}
\caption{\small Detailed performance comparison of SNIP and the E2E Transformer baseline on 400 synthetic validation functions. Performance is measured by \textbf{Mean Accuracy (top)} and \textbf{Mean Expression Complexity (bottom)} across different levels of formula and input difficulties: \textbf{(a)} number of binary operators, \textbf{(b)} number of unary operators, \textbf{(c)} input dimension, and \textbf{(d)} number of input centroids. Mean accuracy reflects the percentage of solutions with $R^2 > 0.99$. Mean expression complexity quantifies the average prefix length of generated expressions. 
% Evaluation of \modelname and E2E Transformer baseline in terms of \textbf{Mean Accuracy (top)} and  \textbf{Mean Expression Complexity (bottom)} on 400 In-domain Synthetic Validation Functions. Mean accuracy is assessed by the proportion of solutions with an $R^2$ score greater than $0.99$, and Mean Expression Complexity is quantified as the average of prefix length in generated expressions.  The analysis is stratified across different levels of formula and input difficulties: \textbf{(a) number of binary operators, (b) number of unary operators, (c) input dimension, and (d) number of input centroids.} 
% Performance comparison of \modelname and E2E transformer baseline in terms of mean accuracy defined as the ratio of solutions with $R^2>0.99$ over $400$ in-domain synthetic validation functions
% across different levels of formula and input difficulties: \textbf{(a) number of binary operators, (b) number of unary operators, (c) input dimension, and (d) number of input centroids.}
}
\label{fig:sr-indomain}
\vspace{-0.5em}
\end{figure}

\paragraph{Detailed Results.}
We evaluate \modelname against the E2E transformer baseline on an in-domain synthetic validation set. This set consists of 400 equation examples following the data generation protocol from Sec.~\ref{sec:app-pretraindata}. Functions uniformly vary across difficulty factors: input dimension $d \sim \mathcal{U}(1,d_{max})$, number of unary operators $u \sim \mathcal{U}(0,u_{max})$, binary operators $b \in \mathcal{U}(d-1,d+b_{max})$ where $d_{max}=10$, $b_{max}=4$, $u_{max}=4$. We generated sequences of equation examples for each function by providing $200$ input points $(x,y)$, and assessed prediction accuracy on another set of $200$ test points. The fitting accuracy, denoted as $Acc (R^2>0.99)$ , is the proportion of solutions where the $R^2$ score exceeds 0.99. Additionally, the complexity is quantified as the number of nodes within the expression tree of the generated equations. Figure \ref{fig:sr-indomain} presents a detailed comparison between \modelname and the E2E transformer baseline regarding fitting accuracy and the complexity of the derived equations on the in-domain datasets. The results demonstrate how the models' performance is affected by increasing formula complexity, as characterized by a higher number of operators and input dimensionality. Results show that as problem difficulty grows via more operators, \modelname maintains higher accuracy with a lower corresponding complexity increase compared to the E2E baseline.

\paragraph{Ablation over Impact of Latent Space Optimization.}
To further demonstrate the impact of SNIP Latent Space Optimization (LSO), we conducted additional experiments using optimizers from the Nevergrad library \citep{nevergrad}. The results on 400 held-out synthetic validation functions are shown in Table \ref{table:lso_ng}. We evaluate four configurations: (1) SNIP without LSO, (2) SNIP with LSO using our employed Grey Wolf Optimizer (GWO), (3) SNIP with LSO using Nevergrad's NGOpt optimizer, and (4) SNIP with LSO using Nevergrad's TwoPointsDE optimizer. 
Table \ref{table:lso_ng} shows the results of experiments, highlighting the mean $R^2$ score greater than $0.99$ and the mean complexity of the discovered equations. These metrics provide insight into the performance of the SNIP model with LSO using different optimization strategies. The results conclusively demonstrate the substantial gains provided by adding LSO to harness SNIP's latent space for symbolic regression. With LSO, the mean accuracy improves from $0.683$ to over $0.80$ regardless of the gradient-free optimization algorithm. The minor differences between optimizers can be attributed to variances in implementation and parameter tuning. Therefore, these additional experiments and analysis help demonstrate the significant benefits unlocked by performing LSO over SNIP's semantic and continuous latent representations. The substantial gains in accuracy underscore the importance of LSO as an integral component of our overall approach in using \modelname for SR.

% \paragraph{\textcolor{blue}{Ablation over \modelname architecture choices.}}

\begin{table}[!ht]
\centering
\renewcommand{\arraystretch}{1.1}
\caption{Performance comparison of \modelname with different LSO configurations.}
\label{table:lso_ng}
\vspace{-0.5em}
\resizebox{0.65\linewidth}{!}{
\begin{tabular}{lcc}
\toprule
Model Configuration & $R^2>0.99$ & Complexity \\
\midrule
SNIP w/o LSO & 0.683 & 28.43 \\
SNIP (LSO w/ GWO) & 0.820 & 29.95 \\
SNIP (LSO w/ NGOpt) & 0.805 & 30.21 \\
SNIP (LSO w/ TwoPointsDE) & 0.805 & 29.91 \\
\bottomrule
\end{tabular}
}
\end{table}

% \subsection{\textcolor{blue}{Additional Ablation Studies}}
% % \vspace{-1.0em}
% \paragraph{\textcolor{blue}{Impact of Latent Space Optimization.}}
% \textcolor{blue}{ Table below ....
% }

\end{document}